\documentclass[10pt]{article}
\usepackage[preprint]{tmlr}

\usepackage{amsmath,amsfonts,bm}

\def\eqref#1{equation~\ref{#1}}

\def\1{\bm{1}}

\DeclareMathAlphabet{\mathsfit}{\encodingdefault}{\sfdefault}{m}{sl}
\SetMathAlphabet{\mathsfit}{bold}{\encodingdefault}{\sfdefault}{bx}{n}

\usepackage{hyperref}
\usepackage{url}
\usepackage{booktabs}
\usepackage{graphicx}

\title{The Review Lottery: Benchmarking an Observational Estimator of Peer-Review Noise (ICLR 2017-2025)}

\author{\name Feilian Huang \\ \addr Independent researcher}

\def\month{10}
\def\year{2026}
\def\openreview{\url{https://openreview.net/forum?id=XXXX}}

\begin{document}
\maketitle

\begin{abstract}
How much would a conference accept/reject decision change if the same paper were reviewed by a different set of reviewers? Only the NeurIPS 2014 and 2021 consistency experiments measure this directly. We build an observational estimator from public review data and benchmark it against those experiments, then apply it to nine years of ICLR (2017--2025; 36,113 papers, 134,912 reviews). An ordered-probit decomposition separates paper quality from reviewer noise ($\rho$ 0.38-0.58); a logistic decision model maps scores to decisions; posterior draws simulate two committees at each year's actual reviewer count. Estimated two-committee disagreement is 20-25\% on the paper's own estimand. Validated against the NeurIPS 2021 experiment on the decided-paper cohort: at $k=4$ the estimator overstates consistency (21.9\% vs 26.6\%, p$\approx$0.008); at $k=3$ it is not rejected (p$\approx$0.15). Adding decision-stage noise (Bernoulli variant) raises $k=4$ disagreement to 25.7\%, inside the anchor CI. We find no robust trend in decision instability: the spliced actual-k series shows +0.25pp/year (p=0.38). At fixed $k=2$, disagreement rises +0.62pp/year (p=0.017) on the decided-paper cohort, but a composition sensitivity including withdrawn papers cuts the slope to +0.21\,pp/year (p=0.39) --- roughly two-thirds of the nominal rise is withdrawal composition, not rising reviewer noise. The review lottery is real. Whether it is growing cannot be separated from withdrawal composition effects in these data. Headline numbers are post-registration exploratory estimates.
\end{abstract}

\section{Introduction}

Peer review at large machine learning conferences rests on an uncomfortable fact: the same paper, sent to a different set of reviewers, might receive a different decision. This paper estimates a single counterfactual quantity: for a paper reviewed in a given year, the probability that replacing its reviewer pool and re-running the complete process --- independent reviews, discussion, rebuttal, area-chair decision --- would change the accept/reject outcome.

The only direct measurements of this counterfactual come from two randomized experiments. At NeurIPS 2014, 166 papers were reviewed by two independent program committees, which disagreed on 43 papers (25.9\%); the fraction of accepted papers retained by the other committee (accept precision) was 50.6\% (44/87; Cortes and Lawrence, arXiv:2109.09774). At NeurIPS 2021, the experiment was repeated at larger scale ($n=882$): the committees disagreed on 203 papers (23.0\%), with accept precision 49.4\% pooled (99 papers accepted by both committees, of 206 and 195 per committee; the paper's reported 50.6\% is the complement, i.e., the flip rate), and committee mean-score correlation 0.586 \citep{beygelzimer2023}. The 2021 paper does not state the per-committee reviewer count directly. We determine it empirically: a third-party mirror of public OpenReview data (AlgorithmicResearchGroup/openreview-papers-with-reviews, accessed 2026-10-06) contains 10,736 NeurIPS 2021 reviews over 2,768 papers --- mean 3.88 reviews per paper, median 4, mode 4, with 79\% of papers receiving four or more reviews. The same mirror reproduces our ICLR 2021 mean of 3.86 reviews per paper exactly, validating the source. The usual reviewer count at NeurIPS 2021 was therefore four, and we validate the estimator at the $k=4$ caliber as primary, with $k=3$ as a sensitivity. These experiments are the field's only ground truth for review noise, and they cannot be repeated casually: each required duplicating the review of hundreds of papers.

This paper asks whether the same counterfactual quantity can be estimated from observational data alone, and how far such an estimator can be trusted. We make two contributions.

First, we construct an estimator of two-committee disagreement from public review scores. A Bayesian ordered-probit variance decomposition separates each paper's latent quality from reviewer noise, year by year, with year-specific cutpoints absorbing scale changes. A logistic decision model maps simulated scores to accept/reject outcomes, approximating the area-chair stage. Drawing from the posterior, we simulate two independent committees of k reviewers for each paper and record the disagreement rate, the probability that an accepted paper flips, and related quantities. The estimator requires only review scores and final decisions, both public on OpenReview.

Second, we validate the estimator against the two experiments before trusting it. The external validation compares simulated disagreement to the reported NeurIPS 2014 and 2021 numbers at matched reviewer-count and acceptance-rate calibers, with the anchors' own sampling uncertainty included in the comparison. The internal check is model-light: on ICLR papers with four or more reviews and known decisions (18,740 papers), we randomly split reviewers into two groups of two, pass each group through the same decision model, and compare the resulting split-half disagreement (200 repetitions) to the $k=2$ simulation. This check assumes no score-generation model, though it shares the decision model with the main simulation.

We apply the estimator to ICLR 2017--2025 and report two sets of findings, corresponding to two research questions in the retrospectively registered analysis plan. RQ-A asks whether the estimator reproduces the two anchors. RQ-B asks how reviewer noise evolved over nine years during which total submissions grew roughly $24$-fold (489 in 2017 to 11,663 in 2025, withdrawals included; 23.9-fold).

Our headline results (\S4.1): the estimator overstates consistency at $k=4$ against the NeurIPS 2021 anchor, with a marginal result at $k=3$. Paper quality accounts for roughly one-third to three-fifths of score variance ($\rho$ 0.38-0.58). Simulated two-committee disagreement is 23-30\% at $k=2$. At fixed committee sizes, disagreement rises about 0.6pp/year over 2017--2025 ($k=2$ passes our robustness rule), roughly two-thirds explained by withdrawal composition; the spliced actual-k series shows no significant trend due to the mechanical 2021 level shift. A post-hoc withdrawn-scores-excluded refit yields a steeper slope, but the conclusion depends on the specification (see Section 4.2 for the truncation-bias caveat) and we do not headline it. The 2020 and 2021 elevations in the accepted-paper flip rate are decision-instability phenomena with different mechanics: 2020 partly reflects a knife-edge interaction between the fitted decision boundary and the coarse scale's discrete grid, plus genuine scale compression; 2021 reflects low signal plus threshold crowding.

The rest of the paper is organized as follows. Section 2 reviews related work. Section 3 describes the estimator and the two validations. Section 4 reports results for the three research questions. Section 5 discusses interpretation and implications. Section 6 states limitations. Section 7 concludes. Appendix A maps each substantive claim to its generating script and output.

\section{Related Work}

\textbf{Randomized audits of peer review.} The NeurIPS 2014 experiment \citep{cortes2021} duplicated review for 166 papers and reported 43 disagreements (25.9\%), accept precision of 50.6\%, and an estimated 50\% subjective share of score variation. The NeurIPS 2021 experiment \citep{beygelzimer2023} scaled the design to 882 papers and found 23.0\% disagreement with 49.4\% accept precision (pooled retained fraction), concluding that inconsistency had not visibly changed since 2014. These are our external validation targets; unlike these experiments, our estimator needs no duplicated review. A broader survey of challenges and computational approaches in peer review is Shah (CACM, 2022).

\textbf{Variance decomposition of ICLR reviews.} The closest prior work is Tran et al.~\citep{tran2020}, who analyzed ICLR 2017--2020 with variance decomposition and simulation, and reported institutional bias and a gender gap. Their analysis has no external validation, no internal anchor, uses Gaussian ANOVA rather than an ordinal model, and ends in 2020. Our work extends the window to 2025, adds dual validation, and handles year-specific rating scales with ordered-probit cutpoints.

\textbf{Process-centric analyses of peer review.} A recent large-scale study of ICLR 2017--2025 (over 28,000 submissions) examines how disagreement, rebuttals, and review dynamics relate to acceptance (Sangkeun Jung, Goun Pyeon, Inbum Heo, Hyungjin Ahn, arXiv:2509.25701). It is descriptive about the disagreement-to-acceptance mapping; it does not decompose variance, simulate counterfactual committees, or validate against experimental anchors. Our contribution is complementary: an estimator of the rerun counterfactual with quantified validation error.

\textbf{Discussion-stage convergence.} Our estimator's counterfactual is the full process including discussion: simulated committees are drawn at the level of the observed, post-discussion scores. A separate, unobserved counterfactual --- what truly independent reviewers would do with no discussion --- is not what we estimate; we discuss it as a limitation in Section 6 rather than claiming our numbers bound it in either direction. Gao et al. (NAACL 2019; arXiv:1903.11367) document conformity bias in a major NLP conference: final scores are largely determined by initial scores plus distance to other reviewers. Kargaran et al. (arXiv:2511.15462) directly measure pre/post-rebuttal ICLR scores and find reviewer disagreement decreases roughly 9-10\% after rebuttal (concentrated in high-quality papers), with initial scores and co-reviewers' scores the strongest predictors of final scores. One consequence is visible in our model: discussion convergence is absorbed into the paper-quality term as if it were signal, which overstates cross-committee consistency; a posterior predictive check finds the model systematically underpredicts the share of zero-within-paper-variance papers (2020: observed 0.2374 vs replicated mean 0.1922 [0.176, 0.209]), the signature of unmodeled within-committee convergence and reviewer heterogeneity. A large randomized trial on discussion herding (Stelmakh et al., arXiv:2011.15083; PLoS ONE 2023) found no evidence that the discussion initiator disproportionately sways outcomes --- the convergence mechanism looks more like mutual adjustment than following the loudest voice --- but that does not contradict the documented post-discussion compression itself. We return to this in Section 6.

\textbf{LLM use in peer review.} Liang et al.~\citep{liang2024} developed a corpus-level maximum-likelihood method for estimating the fraction of review text substantially modified by LLMs, reporting 6.5-16.9\% across ICLR 2024, NeurIPS 2023, CoRL 2023, and EMNLP 2023 (sentence-level; the ICLR 2024 point estimate is 10.6\% $\pm$ 0.2\%, the first affected year). Our analysis plan specified applying their method at the aggregate level; as documented in Section 4.3, their training-data protocol requires a GPT-4-generated reference corpus beyond this project's budget, so this estimation was not feasible and we report only behavioral descriptives, without attribution.

\section{Method}

\subsection{Variance decomposition: ordered probit with annual cutpoints}

For paper i in year y with $k_i$ reviews, reviewer j's score $s_{ij}$ is modeled as an ordered-probit discretization of a latent variable:

$s_{ij}$ ~ OrderedProbit($\theta_i$, $c_y$), $\theta_i$ ~ Normal(0, $\sigma^2_{\mathrm{paper},y}$)

where $c_y$ are year-specific cutpoints. The cutpoints absorb differences in rating scales across years: ICLR used a $1$--$10$ scale in 2017--2019 and 2021, a four-point scale $\{1,3,6,8\}$ in 2020, and a six-point scale $\{1,3,5,6,8,10\}$ in 2022--2025 (verified from the observed score values per year). The latent residual variance is fixed at 1, so the signal-to-noise ratio is

$\rho_y$ = $\sigma^2_{\mathrm{paper},y}$ / ($\sigma^2_{\mathrm{paper},y}$ + 1),

the share of latent score variance attributable to paper quality. The location parameter $\mu_y$ is fixed at 0 for identification (a standard constraint when cutpoints are free, since location shifts are absorbed by the cutpoints).

Scope note on the estimand. The public scores are final scores after rebuttal and discussion (the berenslab mirror stores only final scores; OpenReview overwrites preliminary versions). Reviewers therefore observe each other before scores are finalized (Gao et al., 2019; Kargaran et al., 2025). Our estimand is the full-process counterfactual --- replace the reviewer pool and re-run everything, including discussion and rebuttal --- so the simulated committees operate at the level of these converged scores. Because convergence is absorbed into theta as if it were paper signal, the fitted model attributes too much observed agreement to quality and overstates cross-committee consistency. The direction of this model-side bias is therefore toward understated disagreement; its magnitude is not identified, and we make no bounding claim on the net bias (Section 4.1).

Fitting is done per year with PyMC NUTS (non-centered parameterization, ADVI initialization, tune=1000, draws=1000, target\_accept=0.9) on CPU. All years show zero divergences and split R-hat between 1.00 and 1.01. 2020 and 2021 were refit with 4 chains: 2020 shows R-hat $\le$ 1.005 and ESS 552-2513 on all monitored parameters with $\rho$ unchanged (0.4390 vs 0.4387); 2021 shows R-hat = 1.0022 and ESS = 949 for sigma\_paper with $\rho$ unchanged (0.3761 vs 0.3764), though the lowest cutpoint mixes poorly (R-hat = 1.20, ESS = 16) because only 14 of 11,488 reviews (0.12\%) used the score 1, leaving it weakly identified --- this does not affect $\rho$, the simulations, or any reported conclusion, and is noted as a limitation. A posterior predictive check simulates replicated datasets from the fitted models: the model systematically underpredicts the share of zero-within-paper-variance papers (2020: observed 0.2374 vs replicated mean 0.1922 [0.176, 0.209]; 2021: observed 0.0474 vs replicated mean 0.030 [0.0239, 0.0367]), suggesting unmodeled reviewer heterogeneity and within-committee convergence; this misfit is reported as a limitation.

Data and cohort. The primary data are the public ICLR reviews from the berenslab/iclr-dataset mirror: 36,113 papers and 134,912 reviews for ICLR 2017--2025. Of these, 11,213 papers were accepted, 17,426 rejected, and 7,474 withdrawn (decision unknown); all unknown decisions were verified to be withdrawals, not data corruption. Withdrawn papers with scores enter the variance decomposition (they contribute to theta estimation) but are excluded from decision-model fitting, which requires an observed outcome. The variance decomposition uses papers with at least one scored review (35,617 papers; 496 papers without scores are excluded). Author identifiers (\texttt{author\_ids}, \texttt{authors}) were deleted at load; a column-name scan confirmed no reviewer identity fields, and no cross-paper reviewer linkage is performed, so reviewer stringency is folded into the noise term. One data caveat: the processed 2017 cohort contains zero withdrawn papers (withdrawn\_rate = 0.000), suggesting withdrawals did not enter that year's mirror; longitudinal comparisons are unaffected except where noted.

\textbf{Target population.} The longitudinal analysis targets a precisely defined population: *papers that went through the full decision process* --- i.e., papers for which the conference recorded a final accept or reject decision. This is the only population on which the review-lottery estimand is well defined: our counterfactual replaces the reviewer pool and re-runs the complete process (independent reviews, discussion, rebuttal, area-chair decision), which presupposes that a decision was reached. Withdrawn papers are therefore excluded from the main trend analysis *by design*, not by convenience: they exited before a decision was recorded, so the counterfactual "what if a different committee had reviewed them" has no observable anchor. Three facts follow from this choice. First, the variance decomposition still uses withdrawn papers' scores (they contribute to theta estimation), but the decision model and the disagreement simulation run on the decided cohort only. Second, withdrawals are an endogenous selection --- papers likely to be rejected withdraw --- so their share of the cohort rose from 0\% (2017) to 25.3\% (2025); mixing them in would conflate a composition change with a change in reviewer noise. Third, the main analysis and the withdrawal-inclusive sensitivity below answer different questions: the former estimates how decision instability evolved among decided papers; the latter measures how much of the nominal disagreement trend is composition. We report both.

\textbf{Confidence term (analysis-plan note).} The analysis plan specifies a $\beta\cdot\mathrm{confidence}$ term in the probit mean, but the primary data source (the berenslab mirror) carries no confidence field (0 of 134,912 reviews non-empty), so the main fit omits it. Because the plan was written after fitting (retrospective registration), this is a design fact about the data --- the confidence term cannot be estimated on the primary cohort --- not a deviation from a prospectively locked analysis. A sensitivity analysis refits the same model with $\beta\cdot\mathrm{confidence}$ on the PeerSum ICLR subset (2018--2022, which has confidence; 53,732 reviews): including confidence changes $\rho$ by -0.016 to -0.030 with beta approximately -0.14 to -0.22 (higher-confidence reviewers score slightly lower), a 4-7\% relative shift that does not change the magnitude of the conclusions.

\subsection{Decision model: from scores to accept/reject}

The area-chair stage is approximated by a logistic regression of the binary decision on paper-level score summaries:

$P(\mathrm{accept}\mid\mathrm{mean\_score},\mathrm{sd\_score},\mathrm{year})$ = logistic($\alpha_{\mathrm{year}}$ + $3.37\cdot\mathrm{mean\_score}-0.14\cdot\mathrm{sd\_score}$),

fit on the 28,476 papers with a known decision and scored reviews (5-fold cross-validated AUC 0.933). The simulation intervals in Table 2 propagate two sources of uncertainty: posterior draws of the paper-quality parameters ($B=1{,}000$) and parameter uncertainty in the decision model itself. The coefficient distribution is a nonparametric bootstrap: the 28,476-paper fit cohort is resampled with replacement and the exact sklearn specification (LogisticRegression, C=1.0, max\_iter=2000) refit 1,000 times (0 refits non-converged). Each of the $B=1{,}000$ score-model draws is paired with one coefficient draw --- a single shared coefficient stream across years, so year-to-year comparisons use common random numbers --- and each simulated committee's decision is sigmoid(X $\beta^{(b)}$) $\ge$ 0.5 with that draw's coefficients. Higher mean scores raise acceptance odds; higher within-paper score dispersion lowers them (sd coefficient -0.138, SE 0.027, p < 1e-6), consistent with the descriptive finding that disagreement hurts acceptance ~\citep{acceptance2025}. Year intercepts absorb annual acceptance-rate differences. In simulation, the logistic model is applied as a deterministic 0.5 cutoff on the predicted probability (sigmoid $\ge$ 0.5); no Bernoulli noise is injected at the decision stage, a fact stated explicitly because it matters for the bias-direction discussion in Section 4.1.

Under the deterministic 0.5 cutoff, the logistic rule is itself a threshold rule --- accept iff mean >= (-$\alpha_{\mathrm{year}}$ + 0.1357*sd)/3.3693. It differs from a pure threshold rule (accept iff mean >= a rate-matching threshold) only in threshold placement (fitted intercept vs rate-matching) and a small sd penalty (about 0.04 mean-points per sd point). Agreement between the two rules is therefore a check on threshold placement, not independent corroboration of the estimator. On the 2020 four-point grid the equivalence is nearly exact: the pooled-logistic and threshold-rule decisions coincide on every reachable (mean, sd) cell at $k=2$ and $k=3$, and differ on a single cell at $k=4$ (mean 5.5, sd 2.89, i.e., {3,3,8,8} committees). The threshold rule's single cutoff cannot hit discrete score grids exactly and over-accepts in simulation (e.g., 2017: simulated 0.499 vs actual 0.405), so it remains a coarse robustness check, not a validated alternative. Simulated acceptance rates under the logistic rule track actual rates closely (e.g., 2021: 0.329 actual vs 0.331 simulated at $k=2$).

Because a single slope across three rating scales is a strong assumption (a one-notch move means different things on a 4-point vs 10-point scale), we also fit scale-specific slopes: 3.93 on the 10-point scale (2017-2019, 2021), 2.33 on the 4-point scale (2020), and 3.40 on the 6-point scale (2022--2025), each with 5-fold CV AUC >= 0.949 (computed within each scale subset, hence not directly comparable to the pooled 0.933). The per-scale model is reported as a robustness check; the pooled model remains primary.

One extrapolation deserves explicit discussion: the decision model is fit on committee means computed over the observed review counts ($n=28{,}476$ papers; median 4 reviews, mean 3.77; only 0.4\% of papers have two or fewer scored reviews), but the main simulation applies it to two-reviewer committee means, whose sampling variance is roughly twice (standard deviation about $\sqrt{}$2 times) that of the median fitting-cohort paper. The fitted slopes therefore describe how acceptance responds to a committee mean that has already averaged out more reviewer noise than the $k=2$ inputs they are applied to; the model was never validated at that noise level, and the direction of the resulting bias in disagreement is ambiguous a priori. The internal split-half check (Section 4.1) does not validate this extrapolation: it applies the same decision rule to observed two-reviewer halves, so a mis-specification of the rule at $k=2$ would affect the check and the simulation identically. We flag the $k=2$ application as an unvalidated extrapolation (Section 6).

\subsection{Two-committee simulation}

For each paper with a known decision and scored reviews, the simulation proceeds as follows: (1) draw $\theta_i$ from its posterior; (2) generate two independent sets of k review scores from the fitted ordered-probit model; (3) pass each set through the decision model to obtain two accept/reject decisions; (4) record whether the committees disagree. Posterior draws $B=1{,}000$; k is reported separately for 2, 3, and 4. Summary statistics are the disagreement rate (fraction of papers on which the two committees differ), $P(\mathrm{flip}\mid\mathrm{accepted})$ (fraction of accepted papers the second committee would reject), and $P(\mathrm{flip}\mid\mathrm{rejected})$. The $k=4$ caliber matches the empirically determined NeurIPS 2021 reviewer count for external validation; the $k=2$ caliber matches the internal split-half anchor.

$P(\mathrm{flip}\mid\mathrm{accepted})$ conditions on the first simulated committee's acceptances: among papers the first (reference) committee accepts, the fraction the second committee rejects. The two committees are exchangeable, so which committee is labeled "first" is arbitrary; averaging over both conditioning directions changes every year-k estimate by less than 0.001 (largest difference 0.00096), i.e., Monte Carlo noise only. The actual accept/reject decisions are not used in this quantity.

\subsection{Dual validation}

External anchor (NeurIPS randomized experiments). The reported values are: 2014, $43/166 = 25.9\%$ disagreement (95\% CI [19.2\%, 32.6\%]) and 50.6\% accept precision (44/87 retained; the 49.4\% figure sometimes quoted is the flip rate, 43/87) \citep{cortes2021}; 2021, $203/882 = 23.0\%$ disagreement (95\% CI [20.2\%, 25.8\%]), 49.4\% accept precision pooled (198/401 retained; per-committee retained 48.1\%/50.8\%, whose complements 51.9\%/49.2\% are what the paper quotes), and committee mean-score correlation 0.586 (95\% CI [0.54, 0.63]) \citep{beygelzimer2023}. The 50.6\% figures are easy to confuse: in 2014 it is the precision (retained fraction), in 2021 it is the pooled flip rate (one minus precision). Table 3a lists both experiments' definitions side by side. Because the public NeurIPS data contain only accepted papers, this is a partial validation by numerical comparison. For the primary decided-cohort comparison, simulated acceptance rates are standardized to the decided-paper acceptance rate of 26.2\% (401 accepted of 1,528 decided committee-paper slots) before comparison, since disagreement moves mechanically with the acceptance rate; the anchors' own sampling uncertainty is included --- comparisons are interval-to-interval, not point-to-point. Standardization uses the fitted logistic decision rule with the intercept shifted to hit the target rate; ties (rare at the decided-cohort boundary) are broken independently per committee at random.

Table 3a. Anchor definitions: the 50.6\% in 2014 is precision (retained fraction); the 50.6\% in 2021 is the pooled flip rate (one minus precision).


Internal check (model-light split-half). Among papers with four or more scored reviews and known decisions (18,740 papers, of 23,995 with four or more reviews), we randomly split the reviewers into two groups of two, pass each group's scores through the same decision model, and compute the disagreement rate directly, repeating the random split 200 times. For papers with more than four scored reviews, each repetition re-permutes that paper's reviews, takes the first four in the permutation, and splits them into two groups of two (first two versus next two), so both the four-review subsample and the assignment to halves are re-randomized in every repetition; within-half score standard deviations use ddof=1, and the decision rule applied is the pooled logistic with the year intercept. This check assumes no score-generation model --- it never synthesizes scores --- but it shares the decision model with the main simulation, so we describe it as model-light rather than model-free; it is best read as an in-sample posterior predictive check of the fitted score model plus a validation of the simulation pipeline, not as independent validation evidence (Section 4.1).

\section{Results}

\subsection{RQ-A: the estimator and its validation}

\textbf{Variance decomposition.} Paper quality accounts for roughly one-third to three-fifths of latent score variance in every year (Table 1; Figure 1). Rho is highest in 2017 (0.583 [0.535, 0.629]) and lowest in 2021 (0.376 [0.356, 0.396]), with the remaining years between 0.410 and 0.481. The magnitude is similar to the NeurIPS 2014 variance-decomposition estimate of about 50\% subjective variation, with the caveat that the two quantities live on different scales: $\rho$ is on the latent probit scale and folds reviewer stringency into the noise term (no reviewer identities are available), whereas Cortes and Lawrence's estimate comes from a different decomposition.

\textbf{Full-sample as the primary specification; withdrawal-indicator refit.} The variance decomposition is fit on the full-sample population (Section 3.1) --- all papers' scores, including withdrawn papers' scores, which contribute to theta estimation (Limitation 15 discusses the pre-discussion caveat). The decision model and the disagreement simulation run on the decided cohort only (papers with recorded accept/reject decisions), as specified in Section 3.1. This gives $\rho$ of 0.583, 0.427, 0.481, 0.439, 0.376, 0.474, 0.434, 0.410, and 0.427 for 2017--2025 (0 divergences in all nine years, split R-hat 1.00--1.01; Table 1). Two sensitivities address the withdrawn-score concern. First, a withdrawal-indicator refit: full-sample fit with a withdrawn-paper indicator (delta\_w) on the latent mean, keeping the population prior intact while allowing withdrawn scores their own intercept. The indicator is large and stable (delta\_w $\approx$ $-$1.0 to $-$1.2 latent SD across years), confirming withdrawn papers score systematically lower. The withdrawal-indicator $\rho$ series --- 0.583, 0.406, 0.448, 0.389, 0.329, 0.401, 0.358, 0.330, 0.326 --- sits uniformly below full-sample. This gap is mechanical, not substantive: adding the indicator moves p(1$-$p)$\delta_{\rm w}$$^2$ of between-paper variance --- where p is the withdrawal share --- from sigma$^2$\_paper into the fixed effect. For 2025, p=0.253 and $\delta_{\rm w}$$\approx$$-$1.1 give p(1$-$p)$\delta_{\rm w}$$^2$$\approx$0.23, accounting for nearly all of the full-sample (0.427) to indicator (0.326) $\rho$ gap. Because p rises from 0\% to 25\% over the sample period, the indicator $\rho$ series inherits a mechanical declining slope ($-$0.024/year, p=0.010) that cannot be read as falling signal share. We report the indicator $\rho$ as a conditional estimand, not comparable to the main column. A true heteroscedastic variant --- giving withdrawn scores their own residual variance sigma\_w --- finds sigma\_w $\approx$ 0.92--1.03 in all nine years (2017 unidentified, no withdrawals): withdrawn scores are not noisier than decided scores. The concern in Limitation 15 (pre-discussion scores have higher dispersion) is not borne out; the withdrawn-score difference is in the mean (delta\_w), not the variance. Second, a withdrawn-scores-excluded refit (excluding withdrawn papers' scores) --- undertaken post-hoc, see the deviation note in the header --- gives uniformly lower $\rho$ than full-sample, with the gap generally widening with the withdrawal share, from 0.0pp in 2017 (where the mirror records zero withdrawals, a clean sanity check) to $-$5.6pp in 2025 (with non-monotonic year-to-year variation: e.g., $-$3.5pp in 2021 vs $-$2.3pp in 2022). The withdrawn-scores-excluded series is reported as a post-hoc sensitivity (Table 1, last column); the full-sample series is primary throughout, and the simulation in Table 2 uses full-sample posteriors.

\begin{figure}[t]
\centering
\includegraphics[width=0.85\textwidth]{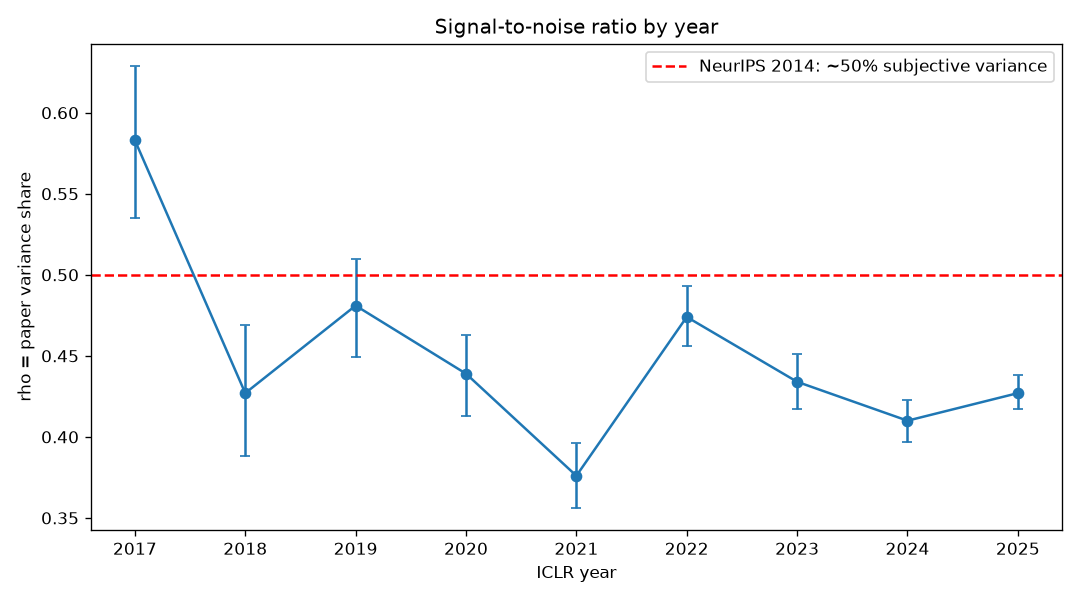}
\caption{Signal-to-noise ratio $\rho_y$ by year (posterior mean, 95\% credible interval). The dashed line marks the NeurIPS 2014 estimate of about 50\% subjective variance.}
\label{fig:rho}
\end{figure}
\begin{table}[t]
\centering
\caption{Signal-to-noise ratio $\rho_y$ by year: posterior mean and 95\% credible interval. Full-sample primary; withdrawal-indicator refit and withdrawn-scores-excluded refit as sensitivities.}\label{tab:t1}
\begin{tabular}{lrrrrrr}\toprule
Year & Papers & Reviews & $\rho$ (full-sample) & 95\% CI & $\rho$, withdrawal-indicator [95\% CI] & $\rho$, withdrawn-scores-excluded [95\% CI] \\
\midrule
2017 & 489 & 1,508 & 0.583 & [0.535, 0.629] & 0.583 [0.532, 0.630] & 0.583 [0.535, 0.629] \\
2018 & 969 & 2,921 & 0.427 & [0.388, 0.469] & 0.406 [0.366, 0.447] & 0.411 [0.368, 0.454] \\
2019 & 1,555 & 4,734 & 0.481 & [0.449, 0.510] & 0.448 [0.416, 0.479] & 0.466 [0.433, 0.496] \\
2020 & 2,561 & 7,775 & 0.439 & [0.413, 0.463] & 0.389 [0.361, 0.416] & 0.416 [0.387, 0.443] \\
2021 & 2,974 & 11,488 & 0.376 & [0.356, 0.396] & 0.329 [0.308, 0.349] & 0.352 [0.329, 0.375] \\
2022 & 3,375 & 13,161 & 0.474 & [0.456, 0.493] & 0.401 [0.381, 0.422] & 0.439 [0.416, 0.461] \\
2023 & 4,919 & 18,575 & 0.434 & [0.417, 0.451] & 0.358 [0.340, 0.375] & 0.411 [0.393, 0.430] \\
2024 & 7,262 & 28,028 & 0.410 & [0.397, 0.423] & 0.330 [0.317, 0.344] & 0.374 [0.358, 0.389] \\
2025 & 11,513 & 46,722 & 0.427 & [0.417, 0.438] & 0.326 [0.314, 0.337] & 0.371 [0.359, 0.384] \\
\bottomrule
\end{tabular}
\end{table}

![Figure 1. Signal-to-noise ratio $\rho$ by year](../revision\_notes/figures/fig14f\_$\rho_{\rm trend}$.png)

\textbf{Two-committee simulation.} Replacing the reviewer pool changes the accept/reject decision for roughly one fifth to one quarter of decided papers on the paper's own estimand (actual reviewer counts per year; Table 2; Figure 2). At $k=2$, disagreement ranges from 23.0\% (2017) to 30.3\% (2025); at $k=3$, from 19.7\% to 26.7\%; at $k=4$, from 17.7\% to 24.2\%. Increasing committee size from 2 to 4 reduces disagreement by 5.3 to 7.7 percentage points across years. Among accepted papers, 31-48\% would be rejected by an independent committee at $k=2$ ($k=4$: 23-40\%), where "accepted" means accepted by the simulated reference committee; this quantity is highest in 2020 (44.7\%) and 2021 (47.9\%) at $k=2$, a pattern examined in Section 4.2. The threshold rule gives similar results in most year-k combinations (all within 2pp except 2021, where it runs higher: $k=2$, 0.281 logistic vs 0.316 threshold under the MAP plug-in); standardizing the acceptance rate removes the 2021 gap (0.260 vs 0.260 at 25.6\% in the fixed-rate rerun), indicating the difference comes from the threshold rule's over-acceptance rather than boundary shape. This agreement is expected given the near-equivalence of the two rules on the score grids (Section 3.2) and is reported as a threshold-placement check, not as independent corroboration.

\begin{table}[t]
\centering
\caption{Simulated two-committee disagreement and accepted-paper flip probability, $k=2/3/4$ (logistic rule, full-sample posteriors). Intervals are 95\% posterior intervals (2.5/97.5 percentiles over joint posterior-draw $\times$ coefficient-bootstrap samples). Point estimates are rounded to three decimals for display; longitudinal trend slopes (\S4.2) are computed from the full-precision posterior-mean series via \texttt{scripts/18\_posterior\_trend.py}, so OLS on the rounded table values reproduces them only approximately.}\label{tab:t2}
\begin{tabular}{lrrrrrrrrr}\toprule
Year & Disagreement $k=2$ & Disagreement $k=3$ & Disagreement $k=4$ & P(flip\ & accepted) $k=2$ & P(flip\ & accepted) $k=3$ & P(flip\ & accepted) $k=4$ \\
\midrule
2017 & 0.230 [0.192, 0.270] & 0.197 [0.164, 0.233] & 0.177 [0.143, 0.211] & 0.307 [0.244, 0.381] & 0.258 [0.198, 0.319] & 0.226 [0.167, 0.283] \\
2018 & 0.293 [0.262, 0.324] & 0.251 [0.224, 0.280] & 0.222 [0.195, 0.250] & 0.369 [0.319, 0.427] & 0.342 [0.291, 0.398] & 0.318 [0.263, 0.376] \\
2019 & 0.270 [0.244, 0.295] & 0.228 [0.202, 0.253] & 0.200 [0.175, 0.223] & 0.369 [0.307, 0.439] & 0.332 [0.279, 0.399] & 0.303 [0.253, 0.356] \\
2020 & 0.273 [0.253, 0.294] & 0.228 [0.210, 0.245] & 0.196 [0.180, 0.211] & 0.447 [0.381, 0.512] & 0.398 [0.359, 0.436] & 0.383 [0.341, 0.425] \\
2021 & 0.281 [0.246, 0.316] & 0.240 [0.219, 0.263] & 0.216 [0.199, 0.235] & 0.479 [0.427, 0.538] & 0.439 [0.399, 0.477] & 0.401 [0.362, 0.440] \\
2022 & 0.282 [0.265, 0.300] & 0.249 [0.233, 0.264] & 0.224 [0.209, 0.240] & 0.352 [0.325, 0.380] & 0.323 [0.295, 0.354] & 0.296 [0.270, 0.323] \\
2023 & 0.292 [0.278, 0.307] & 0.257 [0.244, 0.272] & 0.233 [0.219, 0.247] & 0.376 [0.351, 0.400] & 0.348 [0.323, 0.375] & 0.320 [0.296, 0.344] \\
2024 & 0.299 [0.288, 0.311] & 0.264 [0.252, 0.274] & 0.238 [0.228, 0.250] & 0.405 [0.383, 0.426] & 0.381 [0.360, 0.402] & 0.352 [0.330, 0.374] \\
2025 & 0.303 [0.293, 0.313] & 0.267 [0.258, 0.277] & 0.242 [0.233, 0.251] & 0.384 [0.368, 0.401] & 0.365 [0.348, 0.382] & 0.339 [0.323, 0.356] \\
\bottomrule
\end{tabular}
\end{table}

![Figure 2. Simulated two-committee disagreement vs committee size k](../revision\_notes/figures/fig14f\_disagreement\_vs\_k.png)

\textbf{External validation (decided-on-decided, primary).} The 118 withdrawn papers in the 2021 anchor were withdrawn in both copies --- they enter the 882 denominator but cannot contribute disagreement. The clean comparison restricts to the 764 decided papers: the anchor disagreement is 203/764 = 26.6\% [23.6\%, 29.8\%] (Wilson CI), at the decided-paper acceptance rate of 26.2\% (401/1,528 slots). This matches our estimator's population (the decided cohort) without combining assumptions. On this cohort, simulated 2021 disagreement is 24.0\% [22.5\%, 25.6\%] at $k=3$ and 21.9\% [20.3\%, 23.4\%] at $k=4$. Using interval-to-interval comparison (combining anchor and simulation SEs, per \S3.4): at $k=3$, z$\approx$$-$1.46, p$\approx$0.15 (marginal); at $k=4$, z$\approx$$-$2.63, p$\approx$0.008 --- the estimator overstates consistency at $k=4$. The two independent moments: disagreement (above) and committee correlation --- 0.603 [0.575, 0.630] at $k=3$ (inside anchor [0.54, 0.63]), 0.669 [0.643, 0.693] at $k=4$ (above, no overlap). (Precision is locked to disagreement via precision = 1 $-$ d/(2a): 54.3\% at $k=3$, 58.4\% at $k=4$.) Inverting the test: the 26.6\% anchor disagreement is consistent with an effective pipeline $\rho$ in [0.13, 0.28] (Spearman-Brown mapping calibrated against simulation; see Appendix A), while our fitted 2021 $\rho$ is 0.376 [0.356, 0.396]. The anchor rejects the full pipeline at $k=4$ --- score model, deterministic decision rule, and cross-conference transfer jointly --- not $\rho$ alone.

\textbf{Inclusive-population sensitivity.} For completeness, on the inclusive population (all 882 anchor papers, 22.7\% rate): simulated disagreement is 21.2\% [19.9\%, 22.5\%] at $k=3$ (inside anchor [20.2\%, 25.8\%]) and 19.2\% [17.8\%, 20.6\%] at $k=4$ (point below CI, intervals overlap). The inclusive comparison is less clean: the anchor's withdrawn papers contribute zero disagreement by construction, while simulated ICLR withdrawals pass through the decision model and can disagree. We report it as a sensitivity, not the primary.

Accept precision, correctly defined as the retained fraction (the 50.6\% quoted in the 2021 paper is the flip rate, one minus precision), is 58.4\% at $k=4$ on the decided cohort --- above the anchor's clustering-aware 95\% CI upper edge [43.2\%, 55.2\%] --- against the corrected anchor 49.4\%. At $k=3$, simulated precision is 54.3\%, inside the anchor CI. The clustering-aware CI is a paper-level bootstrap (200,000 resamples of the 882 papers): the 401 acceptance slots belong to 302 distinct papers (99 papers contribute two slots each), so the naive Wald interval [44.5\%, 54.3\%] understates uncertainty. Committee mean-score correlation is 0.669 at $k=4$ on the decided cohort, which does not overlap the anchor CI [0.54, 0.63], against 0.586; at $k=3$ it is 0.603, inside the anchor CI (the anchor CI itself is Fisher-z computed by us from the reported r=0.586, $n=882$; the paper reports no CI). Precision and disagreement are arithmetically locked given the acceptance rate --- the identity precision = 1 - d/(2a) holds exactly on both sides --- so precision is a derived quantity, not an independent second moment to fit.

At $k=3$, the disagreement moment falls inside its anchor CI (24.0\% [22.5\%, 25.6\%] vs [23.6\%, 29.8\%]), so the estimator is not rejected at the $k=3$ caliber. At $k=4$, both moments miss in the same direction --- the simulated committees are too consistent (correlation 0.669 vs [0.54, 0.63], no overlap; disagreement 21.9\% [20.3\%, 23.4\%] vs [23.6\%, 29.8\%], z$\approx$$-$2.63, p$\approx$0.008) --- indicating the model overstates cross-committee consistency at $k=4$. We therefore describe the external exercise as a one-sided consistency check, not a calibration, and we report the k-dependence explicitly: the validation conclusion depends on k, and k itself was chosen post-registration (see the "Two caveats on the k caliber" paragraph below).

Table 3b. External validation summary (decided-on-decided, primary; 2021 at 26.2\% accept rate): $\checkmark$ = point inside anchor CI, ~ = intervals overlap but point outside, $\times$ = no overlap.


Two caveats on the k caliber. First, $k=4$ was chosen post-registration from a third-party mirror of NeurIPS 2021 reviews (mean 3.88 reviews/paper); the mirror's selection rule is undocumented, so the "usual count is 4" inference may be biased upward --- we report $k=3$ throughout as a sensitivity. Second, the committee correlation depends on the within-sample spread of paper quality, so part of the correlation gap may be population mismatch rather than model overstatement.

Via the Spearman-Brown formula, the anchor correlation of 0.586 implies single-reviewer reliability of about 0.26 ($k=4$) or 0.32 ($k=3$); the model's 0.669/0.603 imply about 0.34-0.36 at both k, consistent with the fitted 2021 $\rho$ of 0.376. The discrepancy is concentrated at $k=4$.

The 2021 external validation ran on the 2-chain fit (2 chains $\times$ 1000 draws); the 4-chain refit exists only as convergence diagnostics --- its draws file carries no paper-level z, so it cannot be run through the simulator. The 4-chain refit moves $\rho$ by +0.00026 (0.37636 vs 0.3761), within 3 MCSE of the 2-chain value. The weakly-identified lowest cutpoint (4-chain R-hat 1.20, ESS 16; only 0.12\% of 2021 reviews used score 1): shifting it by the full between-chain disagreement range changes simulated disagreement by a negligible amount (under 0.01\,pp); deleting the score-1 category entirely has no measurable effect. The weak identification is numerically irrelevant to every reported quantity.

\textbf{Bias direction.} The model-side bias plausibly pushes toward understated disagreement: discussion convergence is absorbed into the paper-quality term as if it were signal, so simulated committees are more consistent than real ones. Two lines of evidence are consistent with this direction. First, the simulated committee mean-score correlation exceeds the anchor's at $k=4$. Second, the posterior predictive check underpredicts the share of zero-within-paper-variance papers. However, a third, omitted component works in the same direction: the simulation uses a deterministic 0.5 cutoff, removing all area-chair-stage randomness. The logistic decision model's AUC of 0.933 leaves substantial residual; the AC's idiosyncratic component should be independent across the two committees. A Bernoulli decision-noise variant (sampling decisions from the fitted probabilities rather than thresholding) adds +3.8pp at $k=4$, taking simulated disagreement from 21.9\% to 25.7\% --- inside the decided anchor CI [23.6\%, 29.8\%]. Decision-stage noise alone could therefore close the $k=4$ gap, without invoking discussion convergence. We report the [deterministic, Bernoulli] interval as a sensitivity on the decision-stage component only: the deterministic simulation omits AC noise, the Bernoulli variant treats all residual as independent noise (an upper bound on that component). The magnitude of the model-side bias is not identified, and the remaining components --- cross-conference differences and rate standardization --- have unknown directions. We make no bounding claim on the net bias combining all components.

\textbf{Validation power: negative controls.} The external check has asymmetric, limited power. Forcing $\rho$ to 0.2 (reviewers roughly twice as noisy as fitted) gives 25.6\% [24.1\%, 27.1\%] at $k=4$, inside the anchor CI: the check does \textbf{not} reject a much-noisier-reviewer world. Forcing $\rho$ to 0.7 (much quieter reviewers) gives 11.5\% [10.3\%, 12.6\%] at $k=4$, far below the anchor CI with no interval overlap: cleanly rejected. A Tran et al.-style Gaussian ANOVA score model (continuous Gaussian scores; implied $\rho$ 0.37, near the probit 0.376) gives 20.9\% [19.4\%, 22.3\%] at $k=3$, below the decided anchor CI [23.6\%, 29.8\%]: it does not pass on the decided cohort. The anchor therefore distinguishes the probit from the ANOVA specification on the decided cohort. Had all controls landed inside the CI, the check would have been vacuous --- they did not (the $\rho$ = 0.7 control fails decisively), so the check has real but one-sided power: disagreement saturates as $\rho$ $\to$ 0, which is why noisy counterfactuals are hard to reject.

\textbf{Cross-year robustness.} The 2021 result is not 2021-specific. Standardizing every year's model to the 26.2\% decided accept rate ($k=4$) and comparing each against the NeurIPS 2021 decided anchor [23.6\%, 29.8\%]: 3 of 9 years have their simulated point inside the anchor CI, and 6 of 9 have overlapping intervals. The anchor discriminates less sharply on the decided cohort than on the inclusive cohort, reflecting the wider decided anchor CI.

\textbf{Internal check.} The model-light split-half check (Table 3) runs on the 18,740 papers with four or more scored reviews and known decisions (of 23,995 with four or more reviews). The split-half values are computed on the known-decision cohort (Table 3 reports this cohort's n). Recomputing the $k=2$ simulation on exactly the anchor papers gives, for the bootstrap values in Table 2 vs the anchor-cohort recomputation: 2017: 0.230 $\to$ 0.244; 2018: 0.293 $\to$ 0.318; 2019: 0.270 $\to$ 0.289; 2020: 0.273 $\to$ 0.281; 2022--2025 within 0.004 (2021: 0.281 $\to$ 0.260, a 0.021 gap). No year shows a significant model-anchor difference (all p >= 0.52; smallest p = 0.529 in 2023). The pooled split-half rate is 0.296 [0.291, 0.301] against the model 0.294 [0.292, 0.297], difference -0.002, p = 0.62. The 2018 anchor (n=18) is uninformative and excluded from the agreement count.

\begin{table}[t]
\centering
\caption{Internal check: model-light $2{+}2$ split-half disagreement (paper-level bootstrap CI) versus the $k=2$ simulation on the same papers (posterior 2.5/97.5 percentiles); p = two-sided empirical. 2018 (n=18) marked uninformative.}\label{tab:t3}
\begin{tabular}{lrrrrr}\toprule
Year & n (anchor) & Split-half [95\% CI] & Model $k=2$ [95\% CrI] & Difference & p \\
\midrule
2017 & 44 & 0.284 [0.167, 0.394] & 0.244 [0.193, 0.300] & -0.039 & 0.55 \\
2018 * & 18 & 0.296 [0.147, 0.467] & 0.318 [0.233, 0.406] & +0.022 & 0.85 \\
2019 & 76 & 0.319 [0.241, 0.405] & 0.289 [0.247, 0.332] & -0.030 & 0.54 \\
2020 & 158 & 0.297 [0.236, 0.359] & 0.281 [0.251, 0.310] & -0.016 & 0.62 \\
2021 & 2,074 & 0.264 [0.249, 0.281] & 0.260 [0.252, 0.268] & -0.005 & 0.60 \\
2022 & 2,076 & 0.288 [0.272, 0.303] & 0.283 [0.274, 0.291] & -0.005 & 0.67 \\
2023 & 2,632 & 0.290 [0.277, 0.305] & 0.296 [0.289, 0.303] & +0.006 & 0.53 \\
2024 & 4,335 & 0.300 [0.288, 0.312] & 0.299 [0.294, 0.305] & -0.000 & 0.99 \\
2025 & 7,327 & 0.307 [0.298, 0.316] & 0.305 [0.300, 0.309] & -0.002 & 0.65 \\
Pooled & 18,740 & 0.296 [0.291, 0.301] & 0.294 [0.292, 0.297] & -0.002 & 0.62 \\
\bottomrule
\end{tabular}
\end{table}

What the internal check actually tests. The $2{+}2$ split-half disagreement is a deterministic functional of the observed within-paper score distribution: for a given paper, whether two random halves disagree depends only on how its scores spread across the decision boundary. The ordered-probit model is fitted to these same scores, and $\rho_y$ear absorbs exactly this within-paper dispersion, so a well-fitting model reproduces the split-half rate almost by construction. The comparison is therefore an in-sample posterior predictive check of the fitted score model, not an independent validation of the disagreement estimator. What it does validate is the simulation pipeline: the fitted decision rule applied to simulated 2-reviewer committees behaves the same as the same rule applied to observed 2-reviewer halves. It cannot validate the score-generating model or the decision rule itself, since both are estimated from --- or shared with --- the check data. We report the split-half comparison as a consistency check, not as independent validation evidence. Under this reading, the 2021--2025 agreement (within 0.6pp on 18,444 papers, all p >= 0.52) is the expected outcome for a well-fitting model, and the early-year gaps are sampling noise. Two further qualifications. First, the split-half halves are drawn from the same converged committee --- the same discussion, the same rebuttal --- so the check is blind to discussion-stage convergence by construction. This is exactly why the internal check can pass while the external committee-correlation check fails: the former never sees the convergence that the latter detects. Second, the p-values are over-conservative: all nine yearly p-values are >= 0.52, an event with probability about 0.1\% under a uniform null, because the split-half and model values are computed on the same papers (positively correlated) but treated as independent, and because sampling and posterior uncertainty are mixed. A paired paper-level bootstrap would be the correct inference; we report the conservative version and do not interpret the high p-values as strong confirmation. An exploratory note: early 4+-review papers are enriched in borderline cases, consistent with fourth reviewers being assigned as tiebreakers, which would mechanically inflate split-half disagreement; this is a hypothesis, not a test.

\subsection{RQ-B: longitudinal change, 2017-2025}

\textbf{Trends in decision instability: fixed-k vs spliced series.} The planned main regression (disagreement ~ log submissions + reviews per paper + scale version, $n=9$ years, OLS) finds no predictor significant at p<0.05 for any of the three outcomes (Table 4). For $k=2$ disagreement: log submissions +0.036 (p=0.069), reviews per paper -0.026 (p=0.454), 2020 dummy -0.021 (p=0.442), six-point-scale indicator -0.008 (p=0.736). The predictors are severely collinear (VIF about 5.8 for log submissions against reviews per paper), and with nine observations and five parameters the multivariate design is intrinsically low-powered --- the null carries almost no information about individual predictors.

The bivariate trends tell a clearer story at fixed committee sizes. At $k=2$, the slope is +0.62pp/year (p=0.017; p=0.021 after dropping 2017); at $k=3$, +0.66pp/year; at $k=4$, +0.67pp/year. These slopes are OLS on the full-precision posterior-mean disagreement series (\texttt{scripts/18\_posterior\_trend.py}, output \texttt{revision\_notes/18\_posterior\_trend.csv}); Table 2 rounds to three decimals, so recomputing from the printed table gives +0.61\,pp/year (p=0.019) --- the same conclusion up to rounding. The $k=2$ trend passes our retrospectively specified robustness rule (p<0.05 in full sample and all single-year-dropped refits). Withdrawal composition accounts for roughly two-thirds of the $k=2$ slope: including withdrawn papers, the slope falls from +0.62 to +0.21\,pp/year (Section 4.2). The paper's own estimand --- the disagreement series evaluated at each year's actual reviewer count ($k=3$ for 2017--2020, $k=4$ for 2021--2025) --- splices two series with a mechanical ~2.4pp drop at the 2021 switch (Table 2: 2020 $k=3$ is 0.228, 2021 $k=4$ is 0.216, while 2021 $k=3$ would be 0.240). This splice yields +0.25pp/year (p=0.24; p=0.92 after dropping 2017; Rubin-pooled p=0.38) --- non-significant, but the non-significance is driven by the mechanical level shift, not by absence of an underlying rise. ICLR's 2021 increase from ~3 to ~4 reviews per paper offset the fixed-k rise. The signal share $\rho$ shows no significant trend (slope -0.012/year, p=0.11; p=0.57 after dropping 2017; pooled p=0.089).

\begin{table}[t]
\centering
\caption{Longitudinal regressions (OLS, $n=9$ years, full-sample population). Top: planned multivariate specification --- no predictor significant at p<0.05 (design intrinsically low-powered with 5 parameters). Bottom: bivariate year trends --- fixed-k disagreement trends are positive and (at $k=2$) pass the robustness rule; the spliced actual-k series does not.}\label{tab:t4}
\begin{tabular}{lrrrr}\toprule
Outcome & log(submissions) & Reviews/paper & 2020 dummy (= 4-pt scale) & 6-pt scale (2022-25) \\
\midrule
Disagreement $k=2$ & +0.036, p=0.069 & -0.026, p=0.454 & -0.021, p=0.442 & -0.008, p=0.736 \\
Disagreement $k=3$ & +0.030, p=0.082 & -0.019, p=0.531 & -0.022, p=0.377 & +0.001, p=0.942 \\
$\rho$ & -0.093, p=0.079 & -0.010, p=0.911 & +0.030, p=0.685 & +0.102, p=0.151 \\
\bottomrule
\end{tabular}
\end{table}


A posterior-draw version of the multivariate analysis --- refitting the regression on each of $B=1{,}000$ posterior draws and pooling with Rubin's rules --- propagates annual estimation uncertainty properly. The pooled 95\% intervals are wide because of within-draw variance --- the OLS sampling error of a nine-observation regression, averaged over draws --- which dominates the between-draw component for every coefficient. All pooled multivariate intervals cover zero, consistent with the low power of the five-parameter design.

*Post-hoc sensitivity (deviation from registered plan).* A withdrawn-scores-excluded refit --- excluding withdrawn papers' scores from the variance decomposition --- was undertaken post-hoc. It yields a steeper disagreement slope (+0.78pp/year at $k=2$, p=0.008, robust to single-year deletion) and a declining $\rho$ (-0.017/year, p=0.039). We report this as a post-hoc deviation, not a pre-specified sensitivity, for two reasons. First, the choice was made after seeing results and after OSF submission. Second, the withdrawn-scores-excluded specification is theoretically questionable: withdrawal is selection on observed scores, so fitting only on the selected sample truncates by the dependent variable, biasing the between-paper variance downward and misspecifying the Normal prior (the truncated thetas are not Normal). If withdrawal depends only on observed scores, the full-sample fit has ignorable selection. The gap between the two $\rho$ series generally widens with the withdrawal rate (0 to 5.6pp), consistent with truncation driving the withdrawn-scores-excluded decline. A dedicated simulation study confirms the mechanism: on synthetic data with true $\rho$ held constant at 0.45 across nine years and truncation rising 0\% $\to$ 25\% (dropping the lowest-scored papers, mimicking withdrawal), the withdrawn-scores-excluded fit manufactures a spurious decline of $-$0.029/year (p=0.0002; year-9 minus year-1 = $-$0.22), while the full-sample control correctly recovers the flat truth (+0.001/year, p=0.68) (\texttt{scripts/18\_truncation\_sim.py}, \texttt{revision\_notes/18\_truncation\_sim.csv}). The withdrawn-scores-excluded $\rho$ decline on real ICLR data ($-$0.017/year) is plausibly the same artifact, not evidence of falling reviewer noise. The trend conclusion is therefore specification-dependent; we do not headline it.

We apply a retrospectively specified robustness rule: a trend claim requires p < 0.05 in the full sample *and* p < 0.05 in every single-year-dropped refit. (This rule was written after seeing the data; it is a stated decision rule, not a pre-registered test.) Under it, the fixed-$k=2$ disagreement trend passes on the full-sample specification (+0.62pp/year, p=0.017; drop-2017 p=0.021). The spliced actual-k series (+0.25pp/year, p=0.24) and the $\rho$ series ($-$0.012/year, p=0.11) do not pass. The withdrawn-scores-excluded disagreement trend also passes the mechanical rule --- all nine single-year-dropped refits are significant at p<0.05 for both $k=2$ (max p=0.029) and $k=3$ (max p=0.012) --- but as established above, that specification truncates on the dependent variable, and the truncation simulation shows the same pattern arising mechanically under a flat truth. We therefore do not treat the withdrawn-scores-excluded rule-passing as evidence for a real trend.

\textbf{Terminology: decision instability, not reviewer noise.} The disagreement rate conflates three things: reviewer noise ($\rho$), the density of papers near the decision threshold, and the acceptance rate. A rise in the disagreement rate is therefore a rise in *decision instability*, not necessarily a rise in reviewer noise. The reviewer-noise question is answered by $\rho$, which shows no significant trend (Table 4, $\rho$ row: bivariate p=0.11, p=0.57 after dropping 2017). The decision-instability question is answered by the disagreement series, whose nominal rise does not survive the actual-k specification (Section 4.2). We keep the two quantities linguistically separate throughout.

\textbf{Composition effects.} Two mechanical suspects were tested. First, acceptance rates vary across years (40.5\% in 2017, 30.9\% in 2020, 32.9\% in 2021, 39-42.5\% in 2022--2025; denominators are papers with known decisions), and P(flip|accepted) correlates -0.70 with the acceptance rate at $k=2$. To test whether the 2020/2021 elevations are acceptance-rate artifacts, all years are re-simulated at a common acceptance rate with the following exact procedure. For each posterior draw, each committee, papers are ranked by committee mean (threshold variant) or by that draw's fitted sigmoid $P(\mathrm{accept})$ (logistic variant, using the draw's bootstrap coefficient sample); the top n = round(target $\times$ P) papers are accepted, and the marginal accepted paper's score is that draw's effective cutoff. A quantile cutoff on the score cannot hit the target on these coarse discrete grids (heavy ties at the cutoff), so the rank rule is what achieves the rate exactly --- and ties at the margin must then be broken randomly: within each draw, tied papers are ordered by fresh Uniform(0,1) keys from an independent seeded stream (seed 1404 + year), consumed in committee-generation order and never perturbing the committee scores themselves (those keep the main pipeline's seed-7+year stream, bit-identical to Section 3.3). This random tie-breaking is itself a knife-edge convention: papers exactly at the marginal score are accepted or rejected by coin toss, so fixed-rate disagreement levels depend on the tie-breaking rule (the quantization experiment moves decision-level numbers by several pp across tie-breaking conventions). The fixed-rate simulation is therefore a sensitivity check on the rate confound, not a primary estimator.

What it shows ($k=2$, logistic variant). At the 25.6\% target, 2020 P(flip|accepted) is 0.504 and 2021 is 0.508; at the pooled-mean 39.35\%, 0.390 and 0.401. Against the proper comparators (2018/2019 at the same common rates), a residual 2020/2021 elevation of 2-6pp persists at both targets. Roughly 30\% of the own-rate 2020-vs-2019 gap is the mechanical acceptance-rate effect; the remainder is not.

The 25.6\% rate does not match the anchor's denominator for the Section 4.1 comparison: the NeurIPS 2021 anchor's disagreement is computed over duplicated papers only, whose acceptance rate is 22.7\% (401 accepted of 1,764 committee-paper slots); that comparison uses 22.7\%. Section 4.2 uses 25.6\% for a different purpose: it is a conventional conference-wide reference rate --- the full NeurIPS 2021 conference acceptance rate (2,334/9,122), from the same conference that provides the external anchor --- used solely to put all nine ICLR years on a common footing and remove the acceptance-rate confound (yearly rates range 30.9\%-42.5\%). Any common rate serves this purpose; the residual 2020/2021 elevation over 2018/2019 persists at the pooled ICLR mean rate of 39.35\% as well. The two sections thus use different denominators for different targets, and neither presents 25.6\% as the anchor's denominator.

\textbf{Tie-breaking sensitivity.} The main pipeline needs no tie-breaking (0 exact ties in 4,428,000 simulated $k=2$ 2020 decisions). Wherever a rate-matching rule forces ties to be broken, the two committees must break them independently --- a shared deterministic tie-break induces artificial agreement. At the 2021 $k=3$ boundary only 2.78\% of papers sit on the margin, so the convention barely matters there (0.01\,pp); in the quantization experiment (47.2\% ties at $k=2$) it dominates. Tie-broken quantities are convention-dependent sensitivities, never primary estimates.

Second, withdrawals (decision unknown) rose from 0\% of the 2017 cohort to 25.3\% in 2025, and are disproportionately low-scored; excluding them could mechanically inflate the decision-instability trend. The test reported here puts the 7,141 scored withdrawals into the disagreement simulation population (treated as rejects in true status, judged by the same decision model). Including them lowers 2025 $k=2$ disagreement from 0.303 to 0.269 (-3.4pp), 2022-2024 by 2.6-2.7pp, 2020-2021 by 1.9pp, and leaves 2017 unchanged (no withdrawals in that year's mirror). The 2017-to-2025 endpoint rise shrinks from +7.3pp to +3.9pp when withdrawals are included; on the slope metric, the $k=2$ trend falls from +0.62 to +0.21\,pp/year, i.e., roughly two-thirds of the nominal slope is withdrawal composition, not rising decision instability. The mechanism is dilution --- withdrawn papers average 4.02 (simple mean of raw scores pooled across the three rating scales), deep in reject territory, so both simulated committees agree on reject --- and the effect is smaller than a naive share-times-rate estimate because some withdrawn papers sit near the decision boundary.

Two disclosures qualify this comparison. First, the 2017 mirror contains zero withdrawal records (see Section 3.1), so the inclusive 2017-to-2025 comparison contrasts populations defined on different bases: 2017 is a pure decided-paper cohort while 2025 mixes decided and withdrawn papers. Second, the inclusive trend is fragile to the 2017 point itself: the bivariate disagreement-on-year slope is +0.0021/year (p=0.387) at $k=2$ and +0.0022/year (p=0.247) at $k=3$ over 2017--2025, and drops to essentially zero once 2017 is excluded (-0.0003, p=0.912 at $k=2$). Under the retrospectively specified robustness rule, the inclusive series supports no trend claim at all --- the remaining +3.9pp endpoint rise is a level difference anchored on a single, smallest, withdrawal-free year, not evidence of rising decision instability.

\textbf{The 2020 and 2021 elevations.} $P(\mathrm{flip}\mid\mathrm{accepted})$ at $k=2$ reaches 44.7\% in 2020 and 47.9\% in 2021, well above the 31-41\% range of the other years. These are decision-instability elevations, not reviewer-noise elevations ($\rho$ shows no corresponding pattern), and the two years have different mechanics.

2020 used the four-point scale $\{1,3,6,8\}$, and the compression fingerprints are clear: 23.7\% of papers have zero within-paper score variance (versus 4.7-13.1\% in other years), score entropy is the lowest in the sample (1.27), and cutpoint spacing relative to the paper signal is the widest (1.67 sigma). Part of the gap between the pooled rule's 2020 $k=2$ flip rate and the per-scale rule's is a knife-edge artifact of where the fitted decision boundary lands on the coarse scale's discrete grid, not of the slope per se. With $k=2$, committee means on $\{1,3,6,8\}$ can take only nine values [1.0, 2.0, 3.0, 3.5, 4.5, 5.5, 6.0, 7.0, 8.0] (verified by enumerating all reviewer-score pairs). The pooled rule accepts iff mean >= 5.3592 + 0.0403*sd; the per-scale rule (slope 2.33) accepts iff mean >= 5.4221 + 0.0136*sd. Both boundaries sit in the gap between 4.5 and 5.5 at sd=0, but the pooled boundary rises faster with sd and crosses the reachable point 5.5 exactly at the (3,8) committee cell (sd=3.54; pooled boundary 5.5016 > 5.5, per-scale boundary 5.4701 < 5.5), so the pooled rule rejects (3,8) committees while the per-scale rule accepts them (Figure 3). That single grid cell accounts for the entire MAP plug-in 0.479 $\to$ 0.411 drop at $k=2$: moving from the pooled to the per-scale rule changes 250,621 of 4,428,000 simulated $k=2$ committee decisions (5.66\%), all in that cell, and literally zero decisions at $k=3$ or $k=4$, where both boundaries fall in the same gaps between reachable means (P(flip|accepted) identical to four decimals). A threshold sweep confirms the knife-edge mechanics: moving a pure-mean threshold across the reachable point 5.5 jumps $k=2$ P(flip|accepted) from 0.411 to 0.479 (+6.8pp); the pooled and per-scale rules behave exactly like pure thresholds placed just above and just below 5.5, respectively. Parameter uncertainty qualifies the comparison: the pooled boundary's position at the (3,8) cell has posterior mean 5.5013 with SD 0.0378 score points, so the posterior puts probability 0.485 on accepting that cell --- the boundary is not resolved at the 0.0016-point resolution of this grid crossing. Under the coefficient posterior the pooled 2020 $k=2$ flip rate is 0.447 [0.381, 0.512] (the Table 2 bootstrap value), against the per-scale rule's 0.434 [0.378, 0.507] (per-scale boundary SD 0.077, P(accepts the cell) 0.661): the two rules' 2020 $k=2$ flip rates are statistically indistinguishable once decision-parameter uncertainty is propagated. The plug-in 0.479 $\to$ 0.411 gap is therefore a point-estimate artifact; the bootstrap intervals do not distinguish the rules.

\begin{table}[t]
\centering
\caption{Knife-edge mechanics, 2020: $k=2$ P(flip|accepted) for a pure-mean threshold t placed in each gap between reachable means, against the two fitted rules. The pooled rule behaves like a threshold just above 5.5; the per-scale rule like one just below.}\label{tab:t5}
\begin{tabular}{lrrrrr}\toprule
t in (3.5, 4.5) & t in (4.5, 5.5) & Per-scale rule & t in (5.5, 6.0) & Pooled rule & t in (6.0, 7.0) \\
\midrule
0.251 & 0.411 & 0.411 & 0.479 & 0.479 & 0.581 \\
\bottomrule
\end{tabular}
\end{table}

![Figure 3. 2020 knife-edge](../revision\_notes/figures/fig14f\_2020\_knifeedge.png)

\begin{figure}[t]
\centering
\includegraphics[width=0.95\textwidth]{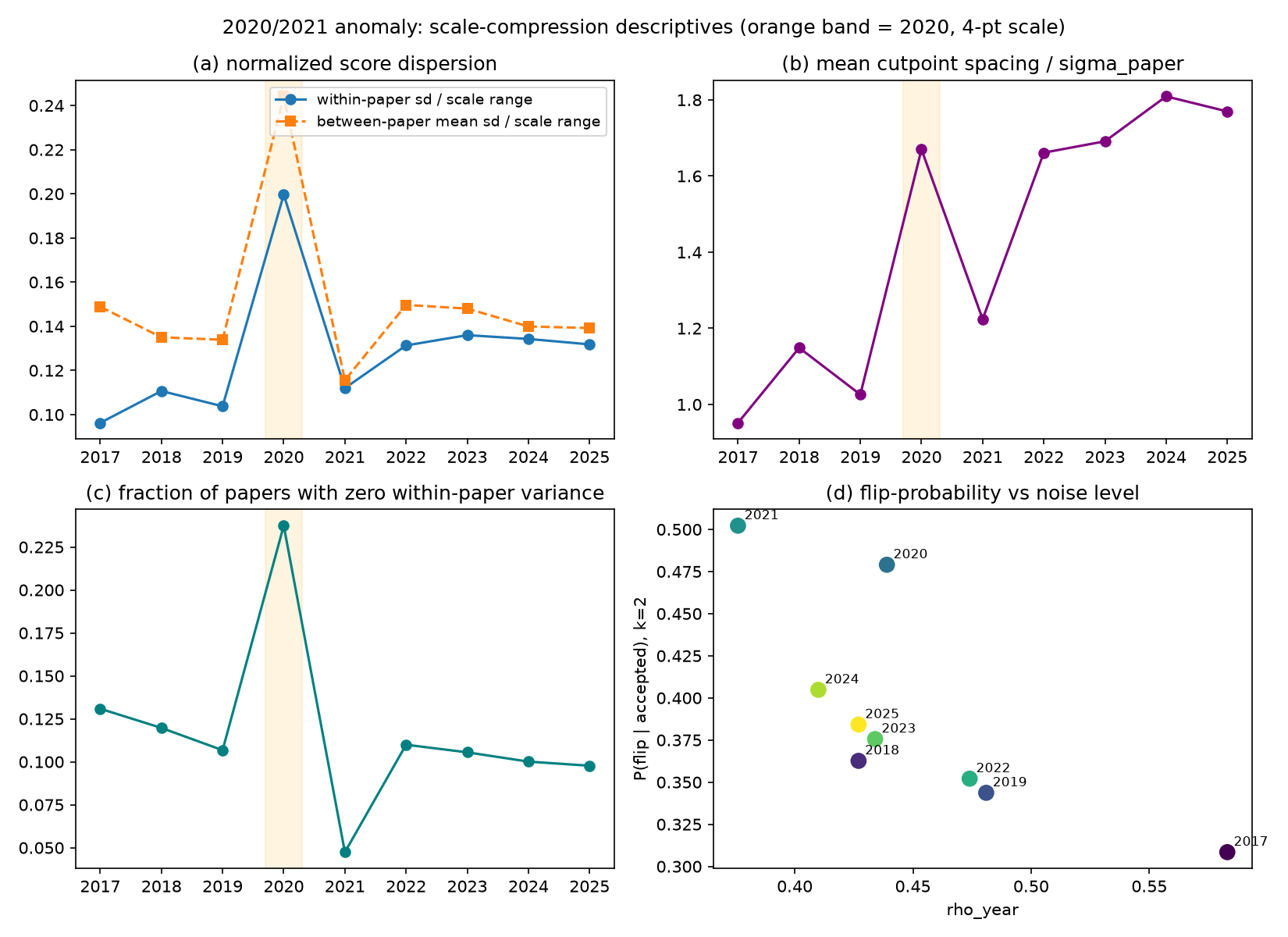}
\caption{2020/2021 anomaly descriptives. (a) Normalized score dispersion; (b) mean cutpoint spacing relative to $\sigma_{\mathrm{paper}}$; (c) fraction of papers with zero within-paper variance (23.7\% in 2020); (d) accepted-paper flip probability against $\rho$. The 2020 four-point scale (shaded) compresses scores; 2021 instead shows low signal with threshold crowding.}
\label{fig:scale-compression}
\end{figure}
2021 shows no compression fingerprint (zero-variance share is the nine-year minimum at 4.7\%). Its mechanism remains low signal plus threshold crowding: $\rho$ is the sample minimum (0.376), and 30.4\% of accepted papers sit within 0.25 of the decision threshold. The fitted 2021 boundary sits 0.0023 score points above the reachable mean 6.0 (the {6,6} cell), i.e., 2021 has its own knife-edge, which is why its interval widens most under parameter uncertainty. The per-scale slope for the 10-point scale (3.93) is close to the pooled slope, so the 2021 result is not a slope artifact.

A uniform knife-edge audit across all years (pooled boundary distance to the nearest reachable $k=2$ grid cell; flagged when the distance is below 0.01 score points and crossing the cell jumps P(flip|accepted) by at least 2pp) flags 2018, 2019, 2020, and 2021; 2017 and 2022--2025 are clean. In 2018 and 2019 the boundary sits 0.006 and 0.007 points from the nearest reachable grid point --- below the boundary posterior SD (0.036 and 0.026) --- so decisions on those cells are effectively coin flips; their point estimates move little under parameter uncertainty (2018: 0.363 $\to$ 0.369; 2019: 0.344 $\to$ 0.369), and the implication is mainly for the intervals. In 2022--2025 the boundary sits 0.19-0.23 points from the nearest grid point, far above the 0.012-0.021 posterior SDs: no knife-edge.

\textbf{Quantization as information loss.} Mapping 2021's ten-point scores to the 2020 caliber $\{1,3,6,8\}$ by nearest neighbor, refitting the ordered probit (zero divergences), and re-simulating isolates the information destroyed by coarsening the scale, holding cohort and year fixed. (This experiment uses the full-sample 2021 specification; the information-loss finding concerns the *change* under coarsening, not the absolute level.) Coarsening to $\{1,3,6,8\}$ is a monotone regrouping of ordinal categories; under a correctly specified ordered-probit model, the latent-scale signal share $\rho$ is theoretically invariant to such coarsening --- information loss should appear as a wider posterior, not a lower point estimate. Empirically the posterior does widen: the $\rho$ 95\% credible interval goes from 0.0396 to 0.0504 in width (27\% wider), and the posterior precision ratio is 0.62 on the credible-interval-width convention ($(0.0396/0.0504)^2$), i.e., roughly two-fifths of posterior precision is lost --- while the point estimate moves from 0.376 [0.356, 0.396] to 0.345 [0.319, 0.369]. The difference posterior for $\rho_{\rm orig}$ - $\rho_{\rm quant}$ is +0.031: the empirical 95\% interval is [+0.0001, +0.0637] but a normal approximation gives [-0.001, +0.063], so the difference is indistinguishable from zero and the data cannot adjudicate whether the point drop is posterior noise or model misspecification. (The two fits are independent MCMC runs; pairing by draw index is display-only.) We therefore claim only the precision-based information loss, which is independent of any decision rule.

\textbf{Scale changes and trend interpretation.} The quantization finding has a direct implication for the longitudinal $\rho$ series: if coarsening the scale can move the $\rho$ point estimate by 0.031 --- the same order as the nine-year trend effects --- then the actual scale changes in 2020 (10-point to 4-point) and 2022 (4-point to 6-point) could themselves induce steps in the $\rho$ series. The 2020 $\rho$ of 0.439 and the 2022 $\rho$ of 0.474 bracket these transitions; we cannot separate a genuine change in reviewer noise from a scale-induced step. The trend analysis in Table 4 includes scale indicators, but with $n=9$ and VIF$\approx$5.8, the design cannot identify scale effects separately from time. We therefore interpret the $\rho$ series as scale-confounded and do not attribute its movements to reviewer noise alone.

A decision-caliber caveat: the $\rho$ result comes from refitting the ordered-probit model on quantized scores and is independent of any decision rule. All disagreement and tie figures in this experiment (47.2\% exact ties at $k=2$, the $\pm$pp deltas under random or deterministic tie-breaking) come from a rate-matched threshold variant (accept iff mean >= 6.0, with random tie-breaking calibrated to the 33.1\% acceptance rate), because the logistic variant's simulated acceptance rate collapses to about 9.8\% on quantized scores and is not comparable. Decision-level numbers from the two pipelines should not be compared directly. At the decision level, the quantized-vs-original disagreement delta is about +7\,pp under independent per-committee random tie-breaking; under the shared deterministic (index) tie-break it ranges from -5.3pp to +3.7pp across $k=2/3/4$ (sign-flipping at $k=2$), because the shared rule induces artificial cross-committee agreement that grows with the tie share. We report the +7\,pp as a convention-dependent sensitivity --- real chairs share tie-breaking information, so it is not a literal causal effect --- and claim only the precision-based information loss as the robust finding.

\subsection{RQ-C: the LLM era (moved to Appendix C)}

The LLM-era analysis (RQ-C) has been moved to Appendix C. In brief: with 2023--2025 review text and confidence data, we find no discontinuity in reviewer confidence around the 2024 breakpoint (flat at 3.6-3.7); adjective occurrence rates rise 7- to 22-fold from the 2023 pre-ChatGPT baseline, a descriptive pattern we do not attribute to AI authorship. The aggregate LLM-share estimation was not feasible within this project's budget.

\section{Discussion}

\textbf{What the validation buys.} The external validation (\S4.1) is a one-sided consistency check: the estimator overstates cross-committee consistency at $k=4$ but is not rejected at $k=3$, and decision-stage noise alone could close the $k=4$ gap. The practical consequence: the "review lottery" can be audited cheaply from public data, but the audit is approximate and inherits the validation's stated uncertainties.

\textbf{Scale design matters, with a caveat.} The quantization experiment's robust finding is about information, not decisions: coarsening 2021's ten-point scores to four widens the $\rho$ posterior by 27\% and cuts posterior precision by about two-fifths (precision ratio 0.62 on the CI-width convention), independent of any decision rule. At the decision level, under the self-consistent independent per-committee random tie-breaking convention, coarsening raises disagreement by about 7pp --- reported as a convention-dependent sensitivity, since real chairs share tie-breaking information. Under a shared deterministic tie-break the effect vanishes or reverses, because the shared rule induces artificial cross-committee agreement that grows with the tie share. The robust claim is information loss, not a decision-level causal effect. The cleaner policy-relevant result is the 2020 descriptive one: the four-point scale compressed 23.7\% of papers to zero within-paper variance, and part of 2020's elevated flip rate is a knife-edge artifact of where the fitted boundary lands on the discrete grid. Conferences choosing coarse rubrics for simplicity should weigh the information cost; the decision-level cost is convention-dependent.

\textbf{On LLMs.} The LLM-era analysis (RQ-C) is reported in Appendix C.

\textbf{The long view.} Over nine years of $24$-fold submission growth, we find no robust trend in decision instability on the paper's own estimand (\S4.2). The fixed-$k=2$ rise does not survive withdrawal composition. We do not claim a growing review lottery.

\section{Limitations}

\begin{enumerate}
\item \textbf{Counterfactual scope.} The estimand is the full-process counterfactual (fresh reviewers, discussion, rebuttal, area-chair decision). The no-discussion counterfactual --- what truly independent reviewers would decide --- is a different, unobserved quantity. Discussion convergence is absorbed into the paper-quality term as if it were signal, which would overstate cross-committee consistency; the simulated committee correlation exceeding the anchor's is consistent with this direction, but its magnitude is not identified, and the cross-conference and rate-standardization components have unknown directions. We make no bounding claim on the net bias in either direction.
\item \textbf{Longitudinal power.} With nine annual observations, the trend tests are weak: the $\rho$ trend test has 80\% power only for slopes of $\pm$0.022/year, and the disagreement trend tests only for $\pm$0.006 to $\pm$0.007/year. Posterior-draw regressions propagate annual uncertainty but cannot create information the nine points lack.
\item \textbf{Collinearity and scale-change steps.} Submission growth, the 2021 shift from ~3.0 to ~3.8 reviews per paper, and scale changes occurred together (VIF about 5.8), so their contributions cannot be separated; the regression attributes nothing robustly to any of them. Separately, the quantization experiment (Section 4.2) shows that coarsening 2021's ten-point scale to the four-point $\{1,3,6,8\}$ moves the $\rho$ point estimate from 0.376 to 0.345 ($\Delta$=0.031, statistically indistinguishable from zero but the same order as the nine-year trend effects). Under a correctly specified ordered-probit model, $\rho$ should be invariant to monotone coarsening; the empirical drop --- whether posterior noise or misspecification --- means the 2020 and 2022 scale changes could themselves induce steps in the $\rho$ series, confounding the longitudinal trend interpretation.
\item \textbf{Decision-model extrapolation.} The decision model is fit on committee means over the observed review counts (median 4 reviews; only 0.4\% of papers have two or fewer scored reviews) but applied to two-reviewer committee means in the $k=2$ simulation; it was never validated at that noise level and the direction of the resulting bias is ambiguous. The split-half check agrees with the $k=2$ simulation within 0.6pp in 2021--2025 where it is tested, but it shares the decision rule with the main simulation and therefore cannot validate the decision rule itself (\S4.1).
\item \textbf{Model misfit.} A posterior predictive check finds the ordered-probit model systematically underpredicts zero-within-paper-variance papers (2020: observed 0.2374 vs replicated 0.1922 [0.176, 0.209]), suggesting unmodeled reviewer heterogeneity and within-committee convergence. The 2021 fit's lowest cutpoint is weakly identified (R-hat = 1.20, ESS = 16; only 0.12\% of reviews used score 1); this does not affect $\rho$ or the simulations.
\item \textbf{Quantization scope.} The quantization experiment runs in one direction only (2021 to the 2020 caliber); its decision-level effect is tie-break-convention-dependent and we claim only the precision-based information-loss result. The difference posterior for $\rho_{\rm orig}$ - $\rho_{\rm quant}$ (+0.031) is indistinguishable from zero (empirical 95\% [+0.0001, +0.0637]; normal approximation [-0.001, +0.063]) and cannot adjudicate posterior noise vs model misspecification. The experiment also extrapolates the decision model to "as if area chairs saw only coarse scores."
\item \textbf{Internal check is in-sample.} The split-half comparison is a posterior predictive check of the fitted score model, not independent validation: a well-fitting model reproduces it almost by construction. Early-year anchor CIs are wide (2018, n=18, is excluded as uninformative).
\item \textbf{Reviewer identities are unavailable.} Reviewer stringency cannot be estimated and is folded into the noise term; the model cannot distinguish a harsh reviewer from a noisy one. This was declared in the analysis plan.
\item \textbf{Partial external validation.} The NeurIPS anchors are numerical comparisons, not refits: the public NeurIPS 2021 reviews predominantly cover accepted papers (the third-party mirror used for reviewer counts contains 2,768 papers vs 2,334 accepted --- the excess are likely rejected papers whose authors voluntarily opened reviews, a self-selected sample), the comparison matches reviewer-count and acceptance-rate calibers rather than re-running the pipeline on NeurIPS data, and cross-conference process differences remain uncontrolled.
\end{enumerate}
\section{Conclusion}

From public review data alone, we estimate that replacing the reviewer pool would flip the accept/reject decision for roughly one fifth to one quarter of decided papers (Table 2). The estimator overstates consistency at $k=4$ against the NeurIPS 2021 anchor but is not rejected at $k=3$ (\S4.1). Over 2017--2025, we find no robust trend in decision instability: the fixed-$k=2$ rise (+0.62pp/year) is indistinguishable from withdrawal composition effects (\S4.2). Coarsening the rating scale destroys measurable information. The review lottery is real; whether it is growing cannot be separated from composition effects in these data.

\section{Appendix A. Claim-evidence mapping}

Each substantive claim in this draft maps to generated evidence. All tables and figures are produced by scripts in the repository; repository-relative paths below.

\begin{table}[t]
\centering
\small
\caption{Claim-to-evidence checklist: each substantive claim maps to generated evidence.}\label{tab:claims}
\begin{tabular}{c p{5.2cm} p{5.2cm} p{3.2cm}}\toprule
\# & Claim & Evidence \\
\midrule
1 & The analysis plan is retrospectively registered; 12*.py through 19*.py analyses are exploratory; withdrawn-scores-excluded refit (15b, 16\_*) is a post-hoc deviation & \texttt{prereg/preregistration\_draft.md} (file created 2026-10-04 23:06 PDT; no version-controlled pre-fitting record); OSF project zgqnw, registration 5vh4e \\
2 & External validation (inclusive population): at the 22.7\% duplicated-paper rate, simulated 2021 disagreement is 21.2\% [19.9\%, 22.5\%] at $k=3$ (inside anchor CI) and 19.2\% [17.8\%, 20.6\%] at $k=4$ (point below anchor CI lower edge 20.2\%, intervals overlap) vs anchor 23.0\% [20.2\%, 25.8\%]; at $k=3$, both moments inside CIs; at $k=4$, correlation misses with no overlap; one-sided power ($\rho$=0.7 rejected, $\rho$=0.2 not) & \texttt{revision\_notes/17\_inclusive\_external.csv}; \texttt{revision\_notes/13a\_anchor\_audit.csv} (anchor values re-verified from arXiv HTML); Beygelzimer et al., arXiv:2306.03262 \\
3 & NeurIPS 2021 reviewer count is 4 (mode/median): third-party OpenReview mirror, 10,736 reviews / 2,768 papers, mean 3.88; same mirror reproduces the ICLR 2021 mean of 3.86 & \texttt{scripts/15a\_neurips2021\_reviewer\_counts.py}; AlgorithmicResearchGroup/openreview-papers-with-reviews (accessed 2026-10-06) \\
4 & Accept precision correctly defined as retained fraction: 2021 anchor 49.4\% with clustering-aware CI [43.2\%, 55.2\%] (paper-level bootstrap; naive Wald [44.5\%, 54.3\%] understates uncertainty); sim 58.4\% at $k=4$ on the decided cohort (above upper edge), 54.3\% at $k=3$ (inside CI); precision = 1 - d/(2a) holds exactly both sides & \texttt{revision\_notes/13a\_anchor\_audit.csv}; \texttt{revision\_notes/14c\_precision\_ci.csv} \\
5 & Model-side bias direction: discussion convergence absorbed into theta plausibly pushes simulated disagreement down (simulated committee correlation 0.669 [0.643, 0.693] at $k=4$ exceeds anchor 0.586 [0.54, 0.63]); magnitude not identified; no bounding claim on net bias & \texttt{revision\_notes/14c\_committee\_re.csv} \\
6 & Rho is 0.38-0.58 across years & \texttt{phase2\_notes/$\rho_{\rm table}$.csv} \\
7 & Two-committee disagreement is 23-30\% ($k=2$), 20-27\% ($k=3$), 18-24\% ($k=4$); accepted-paper flip rate 31-48\% at $k=2$, 23-40\% at $k=4$ (denominator: simulated reference committee's acceptances); intervals propagate score-model posterior draws and decision-coefficient bootstrap draws & \texttt{revision\_notes/14b\_table2.csv} \\
8 & Split-half check agrees with $k=2$ simulation within 0.6pp in 2021--2025 (18,444 papers; all p >= 0.52); pooled 0.296 [0.291, 0.301] vs 0.294 [0.292, 0.297], p = 0.62; reported as in-sample consistency check & \texttt{revision\_notes/13b\_internal\_anchor.csv}; \texttt{revision\_notes/13b\_sampling\_frame.csv} (anchor = 18,740 papers) \\
9 & No robust trend: regression null underpowered (MDE $\pm$0.006-0.007/yr disagreement, $\pm$0.022/yr $\rho$); bivariate $k=2$ trend passes the retrospectively specified robustness rule on the full-sample specification (full p=0.017, drop-2017 p=0.021); actual-k and $\rho$ trends non-significant; inclusive population: no predictor significant in either population & \texttt{revision\_notes/14f\_trend\_rerun.csv}; \texttt{revision\_notes/14f\_trend\_rerun.json} \\
10 & Withdrawal composition: including 7,141 scored withdrawals lowers 2025 $k=2$ disagreement 0.303 -> 0.269 (-3.4pp); 2017--2025 rise shrinks +7.3pp -> +3.9pp; inclusive bivariate slope +0.0021/yr (p=0.387) at $k=2$, +0.0022/yr (p=0.247) at $k=3$; drop-2017 -> -0.0003 (p=0.912); 2017 mirror has zero withdrawal records, so the inclusive comparison mixes populations & \texttt{revision\_notes/13d\_withdrawal\_population.csv}; \texttt{revision\_notes/14e\_inclusive\_trend\_summary.csv}; \texttt{revision\_notes/14e\_dropone\_inclusive.csv} \\
11 & 2020 anomaly: knife-edge --- pooled vs per-scale boundary cross the reachable mean 5.5 at the (3,8) cell; the single cell explains the full MAP plug-in 0.479 -> 0.411 drop at $k=2$ (250,621/4,428,000 decisions, 5.66\%); zero changes at $k=3$/$k=4$; zero exact ties in the main 2020 pipeline; under the coefficient posterior the pooled 2020 $k=2$ flip rate is 0.447 [0.381, 0.512] vs per-scale 0.434 [0.378, 0.507] (indistinguishable); boundary at the (3,8) cell: posterior mean 5.5013, SD 0.038, P(accepts cell) 0.485 & \texttt{revision\_notes/13c\_boundary.json}; \texttt{revision\_notes/13c\_grid\_effect.csv}; \texttt{revision\_notes/13c\_tiebreak\_2020.csv}; \texttt{revision\_notes/14b\_table2.csv}; \texttt{revision\_notes/14b\_2020\_perscale\_interval.json}; \texttt{revision\_notes/14b\_knifeedge\_summary.csv} \\
12 & Quantization: $\rho$ posterior 95\% CI width 0.0396 -> 0.0504 (27\% wider), precision ratio 0.62 (CI-width convention, $(0.0396/0.0504)^2$); difference posterior +0.031 indistinguishable from zero (empirical 95\% [+0.0001, +0.0637]; normal approx [-0.001, +0.063]); no decision-level causal claim & \texttt{revision\_notes/13d\_quantization.csv}; \texttt{revision\_notes/13c\_quant\_audit.md} (decision-caliber audit) \\
13 & 2021 elevation is low $\rho$ (0.376) plus threshold crowding (30.4\% of accepted within 0.25 of threshold); fitted boundary sits 0.0023 points above the reachable mean 6.0 (own knife-edge; interval widens most under parameter uncertainty) & \texttt{phase3\_notes/anomaly\_threshold\_crowding.csv}; \texttt{revision\_notes/14b\_knifeedge\_summary.csv} \\
16 & ITS at 2024 (two post-breakpoint points): no break in within-paper variance (level +0.0096, p=0.938; slope -0.0021, p=0.901); 80\%-power MDE $\pm$0.41 on a 0.017-0.019 baseline; descriptive only & \texttt{revision\_notes/14a\_its\_2024.json} \\
17 & Data provenance (36,113 papers; 134,912 reviews; PII removed; no reviewer identities; final scores) & \texttt{phase1\_report.md}; \texttt{scripts/01\_load\_clean.py} \\
18 & Confidence-term sensitivity: omitting $\beta\cdot\mathrm{conf}$ changes $\rho$ by -0.016 to -0.030 & \texttt{phase2\_notes/conf\_sensitivity.json}; \texttt{scripts/04b\_conf\_sensitivity.py} \\
19 & Sampling diagnostics: 4-chain refits, ESS/Rhat; PPC misfit on zero-variance share (2020: 0.2374 observed vs 0.1922 [0.176, 0.209] replicated; 2021: 0.0474 vs 0.030 [0.0239, 0.0367]) & \texttt{revision\_notes/refit\_4chain.json}; \texttt{revision\_notes/ppc.json} \\
20 & Decision model: sd coefficient -0.138 (SE 0.027, p < 1e-6); deterministic 0.5 cutoff equivalent to threshold rule mean >= (-alpha+0.1357*sd)/3.3693; dual-rule agreement is a threshold-placement check & \texttt{revision\_notes/13e\_sd\_significance.json}; \texttt{revision\_notes/13c\_boundary.json}; \texttt{decision\_model.joblib} \\
21 & Decision-parameter uncertainty: nonparametric bootstrap of logistic coefficients (1,000 refits, 0 non-converged); each posterior draw paired with one coefficient draw (shared stream across years) & \texttt{revision\_notes/14b\_decision\_posterior.json}; \texttt{revision\_notes/14b\_beta\_draws.npz} \\
22 & External-validation power (negative controls): $\rho$=0.2 not rejected ($k=4$: 25.6\% [24.1\%, 27.1\%]); $\rho$=0.7 rejected ($k=4$: 11.5\% [10.3\%, 12.6\%]); Gaussian ANOVA does not pass on the decided cohort ($k=3$: 20.9\% [19.4\%, 22.3\%], below decided anchor CI [23.6\%, 29.8\%]) & \texttt{revision\_notes/14c\_negative\_controls.csv} \\
23 & Cross-year robustness (decided cohort, 26.2\% rate, $k=4$): 3 of 9 years' simulated points inside the anchor CI [23.6\%, 29.8\%], 6 of 9 overlapping intervals & \texttt{revision\_notes/14c\_cross\_year.csv} \\
24 & Committee random effect: exploratory and unidentified (sigma\_c 0$\to$0.3 moves disagreement 22.0\%$\to$23.0\%, correlation 0.618$\to$0.572; anchor first excludes sigma\_c=0.5 via correlation 0.506) & \texttt{revision\_notes/14c\_committee\_re.csv} \\
25 & 2021 validation ran on the 2-chain fit; weakest cutpoint shift has negligible effect on simulated disagreement (<0.01\,pp) & \texttt{revision\_notes/14c\_cutpoint\_perturb.csv} \\
26 & Fixed-rate re-simulation (random tie-break): at 25.6\%, 2020 P(flip\ & accepted) 0.504 [0.471, 0.536], 2021 0.508 [0.477, 0.538]; at 39.35\%, 0.390 [0.365, 0.415] and 0.401 [0.378, 0.424]; residual 2020/2021 elevation of 2.3-6.0pp at 25.6\% and 1.6-5.3pp at 39.35\% over 2018/2019 & \texttt{revision\_notes/14b\_fixed\_rate.csv} \\
27 & Knife-edge audit across years: 2018, 2019, 2020, 2021 flagged (boundary 0.002-0.007 points from nearest grid cell); 2017 and 2022--2025 clean (0.076-0.23 points) & \texttt{revision\_notes/14b\_knifeedge\_summary.csv}; \texttt{revision\_notes/14b\_threshold\_sweep.csv} \\
28 & Target population for the trend analysis: papers with a recorded accept/reject decision (withdrawals excluded by design); total submissions 489$\to$11,663 (23.9-fold); decided papers 489$\to$8,717 (17.8-fold) & \texttt{revision\_notes/14e\_growth\_ratio.json} \\
29 & Tie-breaking sensitivity: 2021 $k=3$ at 22.7\% --- 98.75\% of (draw, committee) pairs have boundary ties (mean 72.2 papers, 2.78\% of cohort, on the marginal score); deterministic index tie-break 0.2198 [0.205, 0.235] vs random 0.2199 [0.204, 0.234] --- disagreement insensitive to the convention; main pipeline needs no tie-breaking (0 exact ties in 4,428,000 simulated $k=2$ 2020 decisions) & \texttt{revision\_notes/15\_tiebreak\_sensitivity.csv}; \texttt{scripts/15\_tiebreak\_sensitivity.py} \\
31 & 2014 anchor at its own 26.2\% accept rate (87/332): sim $k=3$ 0.234 [0.219, 0.251], $k=4$ 0.213 [0.198, 0.230]; both overlap the anchor CI [0.192, 0.326] & \texttt{revision\_notes/15\_anchor2014\_26pct.csv}; \texttt{scripts/15\_anchor2014\_26pct.py} \\
33 & Decided-only $\rho$ sensitivity (post-hoc deviation): refit excluding withdrawn papers' scores (same specification, 0 divergences, R-hat $\le$ 1.017); $\rho$ 0.583/0.411/0.466/0.416/0.352/0.439/0.411/0.374/0.371 for 2017--2025, uniformly $\le$ main (full-sample) series, gap 0.0pp (2017) to $-$5.6pp (2025); series correlation 0.982; see \S4.2 for truncation-bias caveat & \texttt{revision\_notes/15b\_decided\_only\_$\rho$.csv}; \texttt{scripts/15b\_decided\_only\_$\rho$.py} \\
\bottomrule
\end{tabular}
\end{table}

\section{Appendix B. Transparency Statement}

This is an exploratory analysis. The analysis plan was documented after model fitting began (plan file created 2026-10-04, OSF registration submitted 2026-10-05); there is no version-controlled record of the plan predating model fitting. Key choices (robustness rule, $k=4$ validation caliber, decided primary cohort, withdrawn-scores-excluded refit) were made after seeing results and are reported as post-hoc decisions, not confirmatory tests. Deviations from the registered plan: the withdrawn-scores-excluded refit was undertaken post-hoc after OSF submission; the withdrawal-indicator refit was added in response to peer review; the $k=4$ validation caliber was chosen post-registration based on a third-party mirror of NeurIPS 2021 reviewer counts. As a prospective test, we will preregister predictions for ICLR 2026 on OSF before the 2026 data are analyzed.

\section{Appendix C. RQ-C: the LLM era (descriptive)}

The analysis plan specified an interrupted time series for within-paper score variance, review length, and confidence, plus an aggregate LLM-share estimate using the corpus-level method of Liang et al.~\citep{liang2024}. The breakpoint is 2024, not 2023: ChatGPT was released 2022-11-30, while ICLR 2023 reviews were written in October to early November 2022 (submission deadline 2022-09-28; 99.9\% of the 14,335 collected 2023 review notes predate the release, median 2022-10-25). ICLR 2023 is therefore a pre-ChatGPT baseline, and 2024 is the first affected year. With only two post-breakpoint annual points (2024, 2025), the interrupted time series has almost no power --- at 80\% power it can only detect level jumps of $\pm$0.41 on within-paper variance against a 0.017-0.019 baseline (within-paper variance on the standardized score scale) --- and finds no break (level change +0.0096, p=0.938; slope change -0.0021, p=0.901); it is purely descriptive, not evidence of stability.

ICLR 2023--2025 review text and confidence are public on OpenReview. The official OpenReview API presented a bot-detection challenge to automated requests, so the collection used third-party Hugging Face mirrors of public OpenReview data (JerMa88/ICLR\_Peer\_Reviews\_2023; smallari/openreview-iclr2024-peer-reviews-RAW; smallari/openreview-iclr2025-peer-reviews-RAW; accessed 2026-10-05; none declares a license); the attempt is logged in the repository. We collected 89,111 reviews: 2023: 14,335 reviews / 3,792 papers; 2024: 28,028 / 7,262 papers with at least one parsed review (98.1\% of the 7,401 berenslab papers); 2025: 46,748 / 11,520 papers with at least one parsed review (98.8\% of the 11,663 berenslab papers). The 2024 HF mirror counts coincide with the berenslab Table 1 counts; the 2023 mirror covers 76.5\% of berenslab 2023 papers (gap is structural: withdrawn/desk-rejected papers).

What the new data show. Mean confidence is 3.58, 3.64, and 3.67 in 2023--2025, with the low-confidence (<=2) share at 8.4\%, 7.6\%, and 6.9\%: flat, with no visible discontinuity around the 2024 breakpoint. Review length, by contrast, is not comparable across the mirrors, and we draw no length conclusion. The 2023 mirror concatenates more form fields per review than the 2024/2025 mirrors, so the raw means (3,417 vs 2,764 vs 2,960 characters) are a field-concatenation artifact, not a finding: re-extracting 2023 with the same four fields used for 2024/2025 (summary, strengths, weaknesses, questions) gives a mean of 2,565 characters (median 2,255), *below* the 2024 mean of 2,762 (median 2,436) on non-withdrawn papers. We report the re-extracted numbers for transparency and make no claim about length trends. Extending the PeerSum 2018--2022 series (confidence stable at 3.4-3.8), confidence is flat throughout with no discontinuity around 2024.

As pure descriptives with no AI attribution, adjective occurrence rates in the Liang et al. style now read against a genuine pre-ChatGPT baseline. Per-sentence occurrence rates, 2023 (pre; 574,272 sentences) versus pooled 2024-2025 on non-withdrawn papers (1,688,921 sentences; restricting to the eligible sample barely moves the multipliers, from 7.45/21.53/8.65 to 7.37/22.27/8.94): "commendable" rises from 4.7e-05 to 3.5e-04, a 7.4-fold increase [95\% CI 5.1, 10.9]; "meticulous" from 3.5e-06 to 7.8e-05, a 22.3-fold increase [5.3, 86.8]; "intricate" from 3.1e-05 to 2.8e-04, an 8.9-fold increase [5.4, 13.8]. Raw counts put the fragility in plain view: in all of 2023, "meticulous" appears exactly twice in 574,272 sentences (27 "commendable", 18 "intricate"), so the 22-fold multiplier's upper range is fragile --- reported with its wide CI, not as a precise magnitude. The direction and rough magnitude match the pattern Liang et al. report for ICLR 2024 (9.8-, 34.7-, and 11.2-fold increases, Fig. 1), but we make no AI-authorship claim: these are behavioral descriptives only, and per-sentence normalization does not remove the field-composition difference between the 2023 and 2024/2025 mirrors.

The Liang et al. aggregate estimate specified in the analysis plan was not run, and we do not run a reduced ("pilot") version either. Their training-data protocol (Section 3.4) requires generating a reference review with GPT-4 for each paper in the analyzed corpus (about 39,000 in their multi-conference corpus), a paid-API step beyond this project's budget. A small reference corpus would not be a smaller version of the same method: the MLE compares population histograms of adjective occurrence, so an under-powered reference makes the reference distribution too noisy to estimate the AI-share informatively, and the known-alpha validation would have to be redone, i.e., a different, unvalidated method. We also considered the labeled deviation of generating the reference corpus with an open-source model instead of GPT-4, and rejected it on validity grounds, not only cost: the AI reference distribution is model-specific, so swapping models changes the method's identification assumption and voids the Section 3.6 known-alpha validation, which would have to be rebuilt around the substitute model; moreover the reference distribution should approximate the models reviewers actually used in 2024-2025, and the prompt requires paper full text we do not hold. The plan's hard constraints (no training a classifier, no labeling individual reviews) were respected throughout: at no point was any AI-text classifier trained or any individual review labeled. As an external reference point, Liang et al.'s published sentence-level estimate for ICLR 2024, the first affected year, is 10.6\% $\pm$ 0.2\% of review sentences substantially modified by LLMs ~\citep{liang2024}. RQ-C therefore remains descriptive, as the analysis plan specified for the no-clean-breakpoint case.

\begin{figure}[t]
\centering
\includegraphics[width=0.62\textwidth]{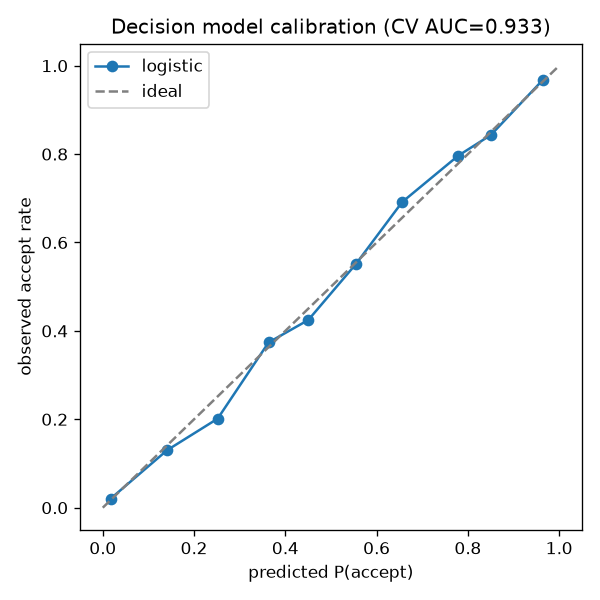}
\caption{Calibration of the logistic decision model: predicted versus observed acceptance rate (5-fold CV AUC $=0.933$). The model tracks the ideal diagonal closely.}
\label{fig:decision-calib}
\end{figure}
\begin{figure}[t]
\centering
\includegraphics[width=0.85\textwidth]{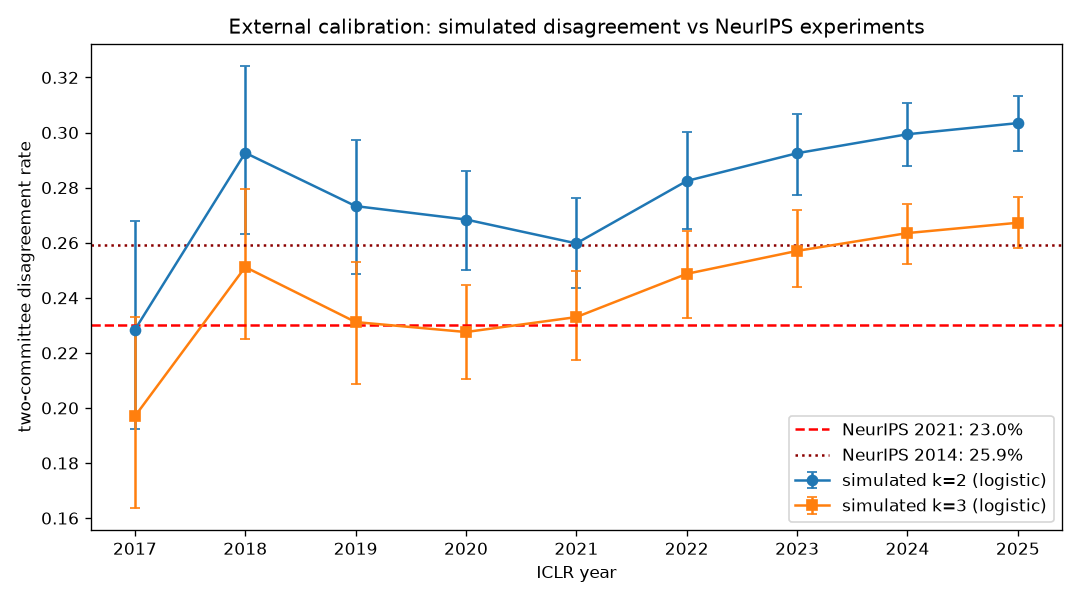}
\caption{External calibration: simulated two-committee disagreement at $k=2$ and $k=3$ against the NeurIPS 2014 (25.9\%) and 2021 (23.0\%) experimental values. At the matched $k=3$ caliber the 2021 simulation sits on the observed value.}
\label{fig:calib-external}
\end{figure}
\begin{figure}[t]
\centering
\includegraphics[width=0.85\textwidth]{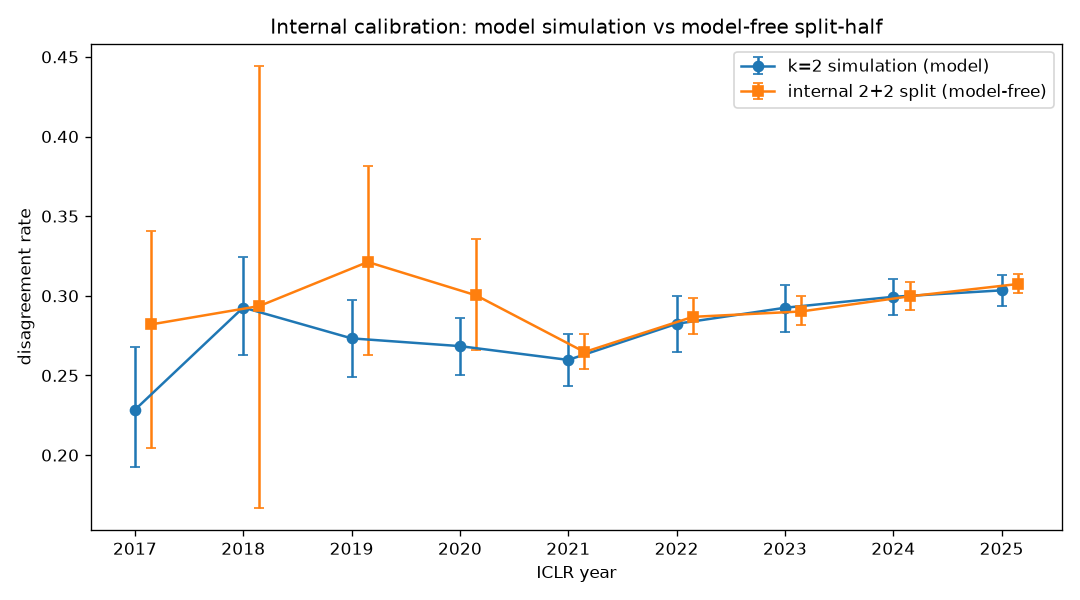}
\caption{Internal calibration: $k=2$ model simulation versus the model-free $2{+}2$ split-half anchor (200 repetitions). Agreement is within 1\,pp from 2021 onward.}
\label{fig:calib-internal}
\end{figure}
\begin{figure}[t]
\centering
\includegraphics[width=0.85\textwidth]{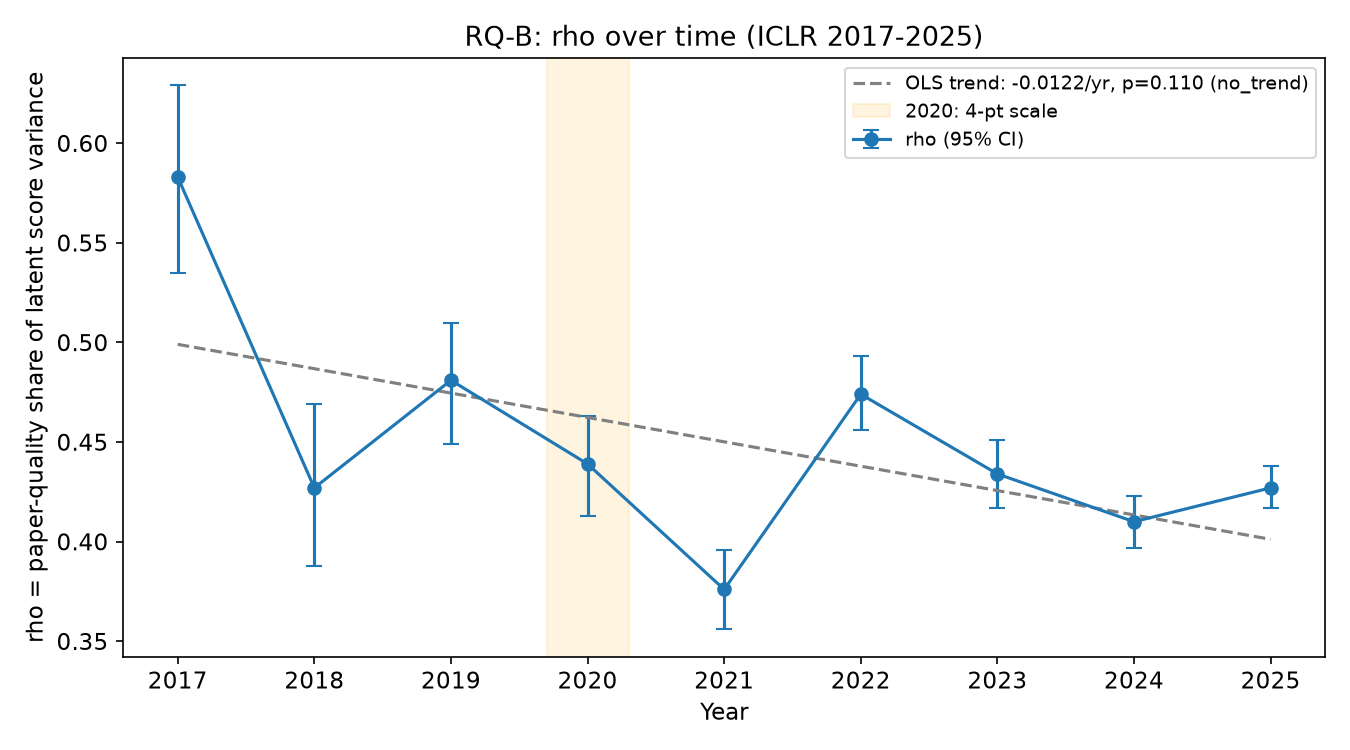}
\caption{RQ-B: $\rho$ over 2017--2025 with the OLS trend ($-0.0122$/year, $p=0.110$; no trend). The shaded band marks the 2020 four-point scale.}
\label{fig:rho-trend}
\end{figure}
\begin{figure}[t]
\centering
\includegraphics[width=0.85\textwidth]{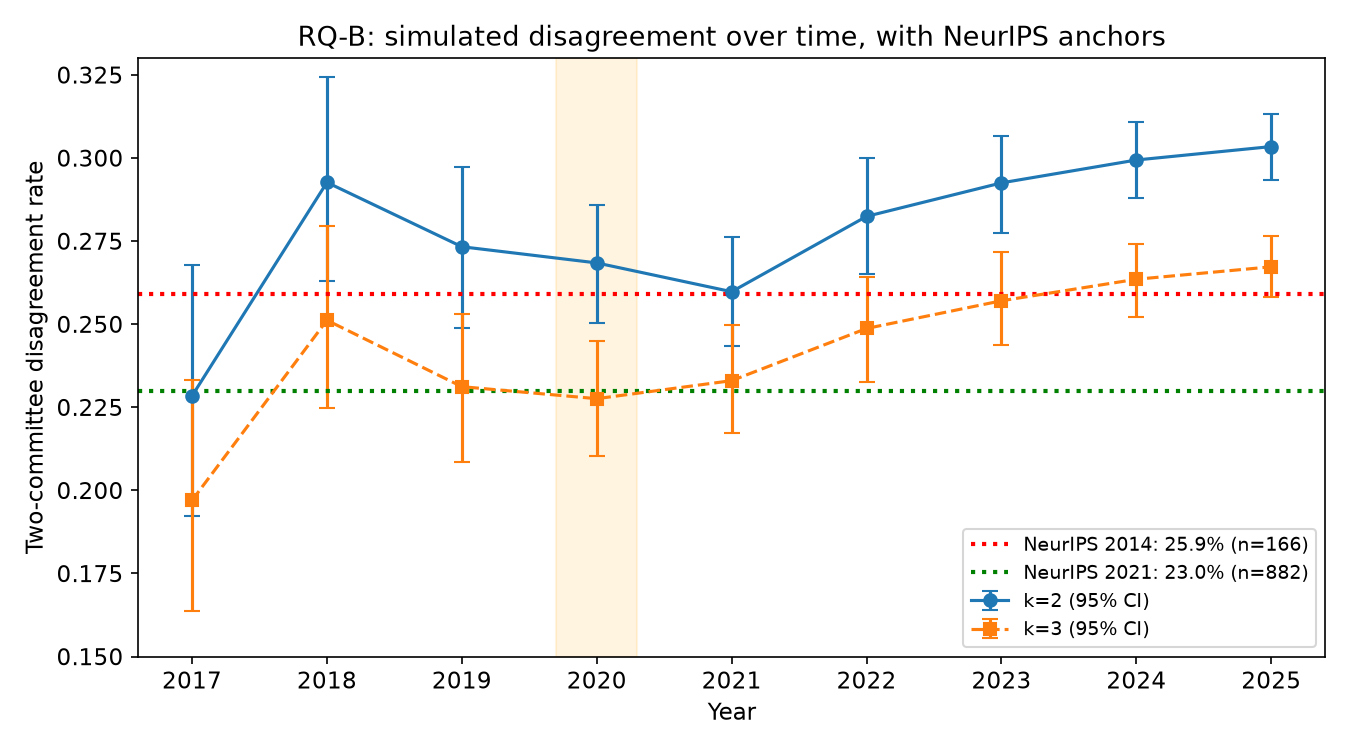}
\caption{RQ-B: simulated disagreement over 2017--2025 at $k=2$ and $k=3$ with the NeurIPS 2014 and 2021 anchors for reference. The nominal rise is not robust to dropping single years.}
\label{fig:disagreement-trend}
\end{figure}
\begin{figure}[t]
\centering
\includegraphics[width=0.95\textwidth]{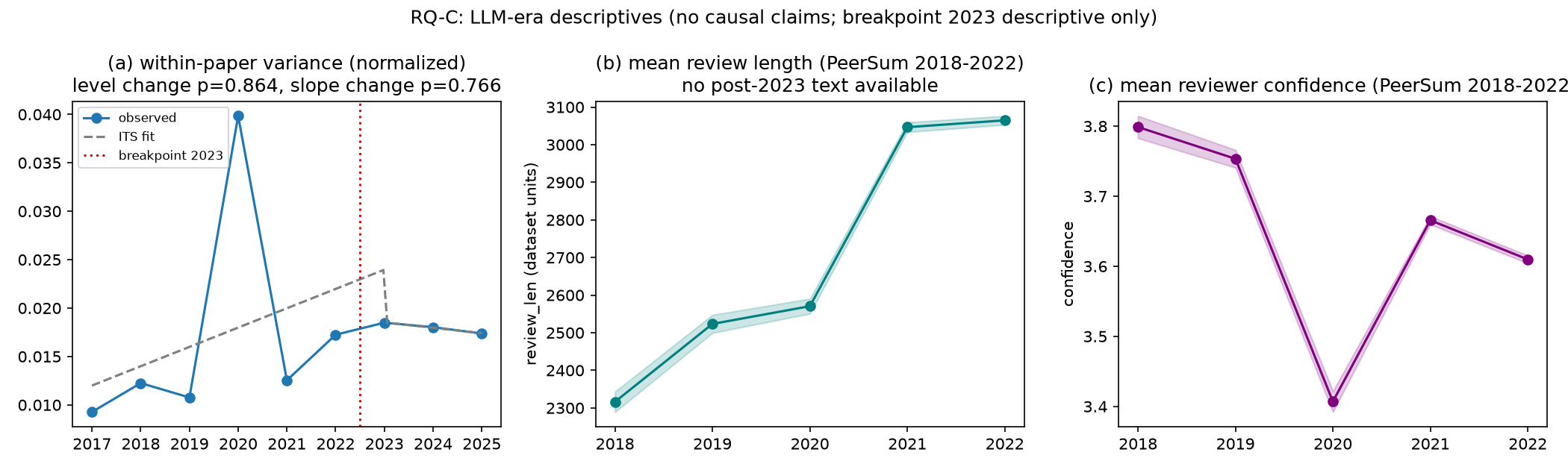}
\caption{RQ-C descriptives (no causal claims). (a) Within-paper score variance with a 2023 breakpoint (ITS: level $p=0.864$, slope $p=0.766$; near-zero power); (b) mean review length, PeerSum 2018--2022 only; (c) mean reviewer confidence, PeerSum 2018--2022 only. No post-2022 review text or confidence data exist.}
\label{fig:llm-descriptives}
\end{figure}

\end{document}